%% file: main.tex
\documentclass[journal]{IEEEtran} 

\usepackage{epsfig}
\usepackage{graphicx}
\usepackage{amsmath}
\usepackage{amssymb}
\usepackage{gensymb}

\usepackage[utf8]{inputenc} 
\usepackage[T1]{fontenc}    
\usepackage{url}            
\usepackage{booktabs}       
\usepackage{amsfonts}       
\usepackage{nicefrac}       
\usepackage{microtype}      
\usepackage{hyperref}
\hypersetup{
    colorlinks=true,
    linkcolor=blue,
}

\usepackage{subcaption}
\usepackage{changepage}
\usepackage{epstopdf}
\usepackage{xcolor}
\usepackage{colortbl}
\definecolor{LightCyan}{rgb}{0.8,1,1}
\usepackage{mathtools}
\usepackage{amsmath}
\usepackage{setspace}
\usepackage{booktabs}
\usepackage{array}
\usepackage{algorithm}
\usepackage{algorithmic}
\usepackage{multirow}
\usepackage{wrapfig}
\usepackage{bbm}
\usepackage{blindtext}

\usepackage{float}

\usepackage{pifont}
\newcommand{\cmark}{\ding{51}}
\newcommand{\xmark}{\ding{55}}

\newcommand{\colvec}[1]{\begin{bmatrix}#1\end{bmatrix}}

\newcommand{\eps}{\epsilon}

\def\bal#1\eal{\begin{align*}#1\end{align*}}
\newcommand{\R}{\mathbb{R}}
\newcommand{\T}{\intercal}

\newcommand{\removed}[1]{}

\newcommand{\revisiontext}[1]{\textcolor{black}{#1}}

\newcommand{\presec}{\vspace*{0mm}}
\newcommand{\postsec}{\vspace*{0mm}}

\title{Unseen Object Instance Segmentation for Robotic Environments}

\author{Christopher Xie, Yu Xiang, Arsalan Mousavian, Dieter Fox
\thanks{C. Xie is with the School of Computer Science and Engineering, University of Washington, Seattle, WA 98195 USA (email: chrisxie@cs.washington.edu).}
\thanks{Y. Xiang and A. Mousavian are with NVIDIA, 2788 San Tomas Expressway
Santa Clara, CA 95051 USA (email: yux@nvidia.com; amousavian@nvidia.com)}
\thanks{D. Fox is with the School of Computer Science and Engineering, University of Washington, Seattle, WA 98195, and also with NVIDIA, 2788 San Tomas Expressway
Santa Clara, CA 95051 USA (email: fox@cs.washington.edu)}
}

\begin{document}
\maketitle

\begin{abstract}
In order to function in unstructured environments, robots need the ability to recognize unseen objects. We take a step in this direction by tackling the problem of segmenting unseen object instances in tabletop environments. However, the type of large-scale real-world dataset required for this task typically does not exist for most robotic settings, which motivates the use of synthetic data. Our proposed method, UOIS-Net, separately leverages synthetic RGB and synthetic depth for unseen object instance segmentation. UOIS-Net is comprised of two stages: first, it operates only on depth to produce object instance center votes in 2D or 3D and assembles them into rough initial masks. Secondly, these initial masks are refined using RGB. Surprisingly, our framework is able to learn from synthetic RGB-D data where the RGB is non-photorealistic. To train our method, we introduce a large-scale synthetic dataset of random objects on tabletops. We show that our method can produce sharp and accurate segmentation masks, outperforming state-of-the-art methods on unseen object instance segmentation. We also show that our method can segment unseen objects for robot grasping.
\end{abstract}


\input{introduction}
\input{related_work}

\input{method_overview}

\input{dsn}
\input{imp}
\input{rrn}
\input{tod}
\input{experiments}

\input{conclusion}

\bibliographystyle{ieeetran}
\bibliography{references}  

\removed{
The first paragraph may contain a place and/or date of birth (list place, then date). Next, the author’s educational background is listed. The degrees should be listed with type of degree in what field, which institution, city, state, and country, and year the degree was earned. The author’s major field of study should be lower-cased. 

The second paragraph uses the pronoun of the person (he or she) and not the author’s last name. It lists military and work experience, including summer and fellowship jobs. Job titles are capitalized. The current job must have a location; previous positions may be listed without one. Information concerning previous publications may be included. Try not to list more than three books or published articles. The format for listing publishers of a book within the biography is: title of book (publisher name, year) similar to a reference. Current and previous research interests end the paragraph. 

The third paragraph begins with the author’s title and last name (e.g., Dr. Smith, Prof. Jones, Mr. Kajor, Ms. Hunter). List any memberships in professional societies other than the IEEE. Finally, list any awards and work for IEEE committees and publications. If a photograph is provided, it should be of good quality, and professional-looking. Following are two examples of an author’s biography.
}

\begin{IEEEbiography}[{\includegraphics[width=1in,height=1.25in,clip,keepaspectratio]{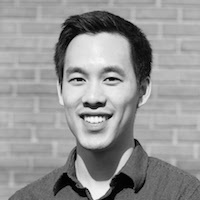}}]{Christopher Xie}
received the B.sc. degree in electrical engineering and computer science from University of California, Berkeley in 2015, and the M.Sc. in computer science from University of Washington in 2017.

He is currently a PhD student in the Paul G. Allen School of Computer Science and Engineering at the University of Washington. From 2016 to 2019, he was a National Defense Science and Engineering Graduate (NDSEG) Fellow. His research interests include applying machine learning to solve robot perception problems. 
\end{IEEEbiography}

\begin{IEEEbiography}[{\includegraphics[width=1in,height=1.25in,clip,keepaspectratio]{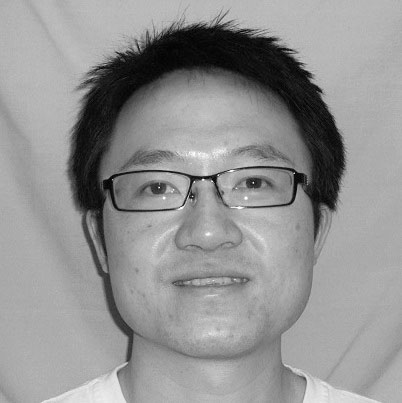}}]{Yu Xiang}
Yu Xiang is a Senior Research Scientist at NVIDIA. He received his Ph.D. in electrical engineering from the University of Michigan at Ann Arbor in 2016. He was a postdoctoral researcher at Stanford University and at the University of Washington from 2016 to 2017, and was a visiting student researcher in the artificial intelligence lab at Stanford University from 2013 to 2016. He received M.S. degree in computer science from Fudan University in 2010 and B.S. degree in computer science from Fudan University in 2007. His research interests focus on robotics and computer vision. His work studies how can a robot understand its 3D environment from sensing and accomplish tasks in the physical world.
\end{IEEEbiography}

\begin{IEEEbiography}[{\includegraphics[width=1in,height=1.25in,clip,keepaspectratio]{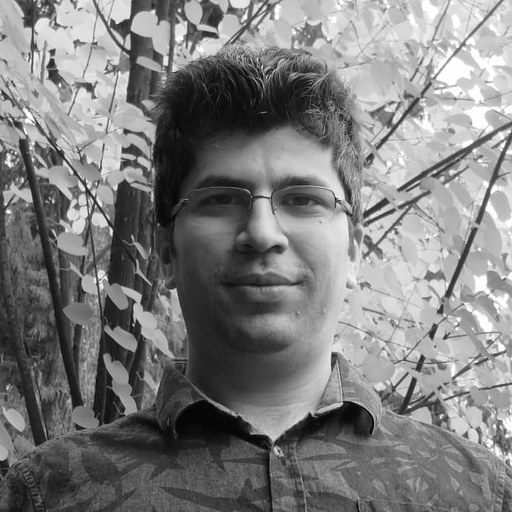}}]{Arsalan Mousavian}
Arsalan Mousavian is a Senior Research Scientist at NVIDIA. He received his PhD in computer science from George Mason University in 2018. Prior to that, he received his M.Sc. degree from University of Tehran in 2013 and his B.Sc. degree from Iran University of Science and Technology in 2010. Arsalan’s research interests are in 3D perception methods that help robots accomplish robot manipulation tasks in the real world. 
\end{IEEEbiography}

\begin{IEEEbiography}[{\includegraphics[width=1in,height=1.25in,clip,keepaspectratio]{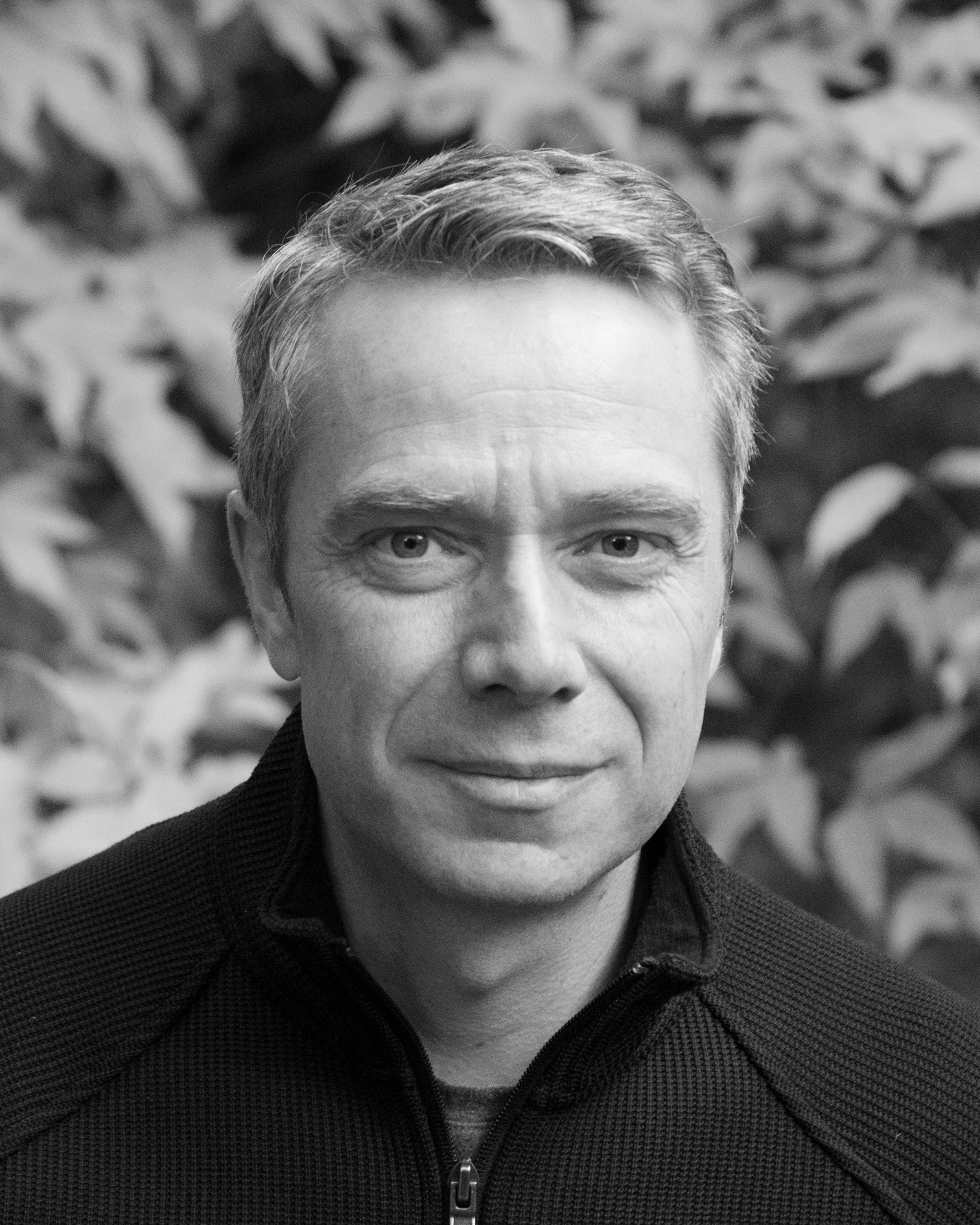}}]{Dieter Fox}
Dieter Fox received the M.S. and PhD degrees from the University of Bonn, Germany, in 1993 and 1998, respectively. 

He is a professor in the Paul G. Allen School of Computer Science \& Engineering at the University of Washington, where he heads the UW Robotics and State Estimation Lab. He is also Senior Director of Robotics Research at NVIDIA.  His research is in robotics and artificial intelligence, with a focus on state estimation and perception applied to problems such as mapping, object detection and tracking, manipulation, and activity recognition. He has published more than 200 technical papers and is co-author of the textbook "Probabilistic Robotics". He is a Fellow of the IEEE and the AAAI, and is recipient of the IEEE RAS Pioneer Award.  He was an editor of the IEEE Transactions on Robotics, program co-chair of the 2008 AAAI Conference on Artificial Intelligence, and program chair of the 2013 Robotics: Science and Systems conference.
\end{IEEEbiography}

\end{document}

%% file: introduction.tex
\section{Introduction}

For a robot to work in an unstructured environment, it must have the ability to recognize new objects that have not been seen before. Assuming every object in the environment has been modeled is infeasible and impractical. Recognizing unseen objects is a challenging perception task since the robot needs to learn the concept of ``objects'' and generalize it to unseen objects. Building such a robust object recognition module is valuable for robots interacting with objects, such as picking up unseen objects or learning to use new tools~\cite{mousavian2019grasp,murali2020clutteredgrasping,mitash2020task}. A common environment in which manipulation tasks take place is on tabletops. Thus, we approach this by focusing on the problem of Unseen Object Instance Segmentation (UOIS), where the goal is to segment every arbitrary (and potentially unseen) object instance, in tabletop environments. 

In order to ensure the generalization capability of the module to recognize unseen objects, we need to learn from data that contains large amounts of various objects. However large-scale datasets with this property do not exist. Since collecting a large dataset with manual annotations is expensive and time-consuming, it is appealing to utilize synthetic data for training, such as using the ShapeNet repository which contains thousands of 3D objects \cite{shapenet2015}. However, there exists a domain gap between synthetic data and real-world data as many simulators do not provide realistic-looking images. Training directly on such synthetic data only usually does not work well in the real world \cite{zhang2015towards}, and synthesizing photo-realistic images with physics-based rendering can be computationally expensive~\cite{hodan2019photorealistic}, making large photorealistic synthetic datasets impractical to obtain.

\begin{figure}[t]
\begin{center}
\includegraphics[width=\linewidth]{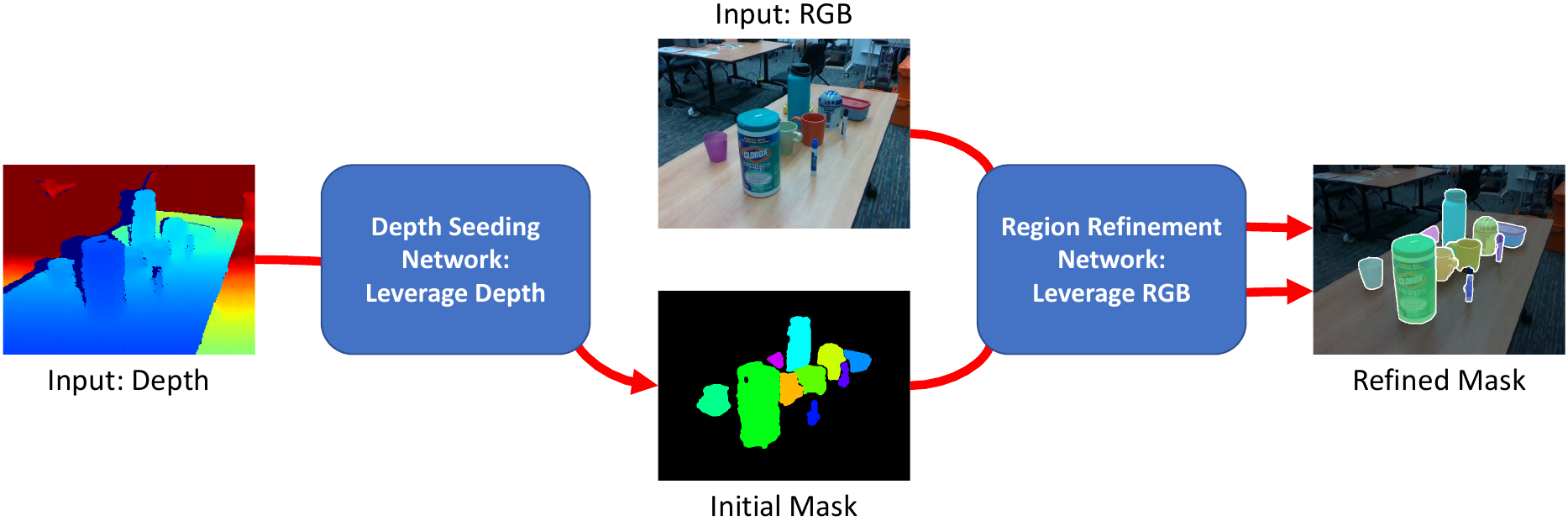}
\caption{High level overview of our proposed two-stage framework. The first stage leverages depth only to produce rough initial masks. The second stage then leverages RGB to refine the initial masks to produce accurate, sharp instance masks.}
\label{fig:teaser}
\end{center}
\end{figure}

Consequently, recent efforts in robot perception have been devoted to the problem of Sim-to-Real, where the goal is to transfer capabilities learned in simulation to real-world settings. For instance, some works have used domain adaptation techniques to bridge the gap when unlabeled real data is available \cite{tzeng2015adapting, bousmalis2018using}. Domain randomization \cite{tobin2017domain} was proposed to diversify the rendering of synthetic data for training. 
While these techniques attempt to fix the discrepancy between synthetic and real-world RGB, models trained with synthetic depth have been shown to generalize reasonably well for simple settings such as bin-picking \cite{mahler2017dex, danielczuk2018segmenting}. However, in more complex settings, noisy depth sensors can limit the application of such methods and models trained on RGB have been shown to produce accurate masks \cite{he2017mask}. An ideal method should combine the generalization capability of training on synthetic depth and the ability to produce sharp masks by utilizing RGB.


In this work, we investigate how to utilize synthetic RGB-D images for UOIS in tabletop environments. We show that simply combining synthetic RGB images and synthetic depth images as inputs does not generalize well to the real world. To tackle this problem, we propose a two-stage network architecture called UOIS-Net that separately leverages the strengths of RGB and depth for UOIS. Our first stage is a Depth Seeding Network (DSN) that utilizes only depth to produce object instance center votes, which are then used to compute rough initial instance masks. We compare multiple architectures for the DSN that produce center votes in 2D and 3D. Training the DSN with depth images allows for better generalization to real-world data. However, the initial masks from the DSN may contain inaccurate object boundaries due to depth senor noise. In this case, exploiting textures in RGB images can significantly help. 

Thus, our second stage is a Region Refinement Network (RRN) that takes an initial mask from the DSN and an RGB image as input and outputs a refined mask. Our surprising result is that, conditioned on initial masks, our RRN can be trained on non-photorealistic synthetic RGB images without adopting any of the afore-mentioned Sim-to-Real solutions. We posit that mask refinement is an easier problem than directly using RGB as input to produce masks, mainly because the mask refinement uses a local image patch as input and focuses on a single object. We empirically show robust generalization across many different objects in cluttered real-world data. In fact, our RRN works almost as well as if it were trained on real data. Our framework produces sharp and accurate masks even when the depth images are noisy. We show that it outperforms state-of-the-art methods including Mask R-CNN~\cite{he2017mask} \revisiontext{and PointGroup~\cite{jiang2020pointgroup}}. Figure \ref{fig:teaser} illustrates our two-stage framework.


To train our method, we introduce a synthetic dataset of tabletop objects in home environments, \revisiontext{which we name Tabletop Object Dataset (TOD)}. Our dataset consists of indoor scenes of random ShapeNet \cite{shapenet2015} objects on random ShapeNet tables. We use the \revisiontext{PyBullet} physics simulator \cite{coumans2019} to generate the scenes and render depth and non-photorealistic RGB. Training our proposed method on this dataset results in state-of-the-art results on multiple \revisiontext{real-world} datasets for UOIS.

We extend our previous work, \revisiontext{UOIS-Net-2D \cite{xie2019uois},} by introducing a novel DSN architecture that reasons in 3D space. As we show in Section~\ref{sec:experiments}, this architecture solves many limitations of producing these center votes in 2D (our previous work), thus providing stronger performance. Additionally, we propose a novel loss function that encourages separation of the center vote clusters and show that this loss function is crucial to achieving strong performance in cluttered environments.


\postsec

%% file: related_work.tex
\section{Related Works}
\presec

\subsection{Category-level Object Segmentation}

2D semantic segmentation involves assigning pixels in an image to a set of known classes. Deep learning has emerged as the most popular tool for solving this problem~\cite{shelhamer2017fully, chen2017deeplab, badrinarayanan2017segnet, chen2018encoder, lin2017refinenet}. \cite{shelhamer2017fully} first introduced the concept of using a fully convolutional architecture (FCN). \cite{chen2017deeplab} designed an architecture that utilizes dilated convolutions in order to increase the receptive field. \cite{badrinarayanan2017segnet, chen2018encoder} further improves performance by introducing decoder architectures on top of the encoders. \cite{lin2017refinenet} proposed a multi-path refinement network with long-range residual connections to enable high-resolution predictions. These methods have demonstrated strong performance on datasets such as PASCAL~\cite{lin2017focal} and COCO~\cite{lin2014microsoft}.

Much work has been devoted to solving the semantic segmentation problem in 3D as well. A common representation of 3D space is voxels; however operating with voxel grids as input can be expensive both computationally and memory-wise. Thus, \cite{graham20183d} introduced a submanifold sparse convolutional operator to preserve spatial sparsity of the input. \cite{choy20194d} further generalized these sparse convolutions to arbitrary kernel shapes, improving performance. Other 3D methods utilize point clouds. \cite{qi2017pointnet} proposed PointNet, a permutation-invariant network architecture to handle point clouds, and \cite{qi2017pointnet++} extended this to a hierarchical network that recursively applies PointNet in order to obtain multi-resolution features, similar to deep convolutional networks with decreasing resolutions via strides/max pooling.

The advent of RGB-D sensors such as Kinect allowed the research community to utilize of both modalities for semantic segmentation, and drove the creation of datasets such as \cite{Silberman:ECCV12}. \cite{ren2012rgb} investigated a combination of kernel descriptors, support vector machines, and Markov random fields for indoor scene segmentation. Both \cite{lai2014unsupervised} and \cite{xiang2017darnn} leverage RGB-D videos, extracting 2D features from each RGB frame and integrating them with a reconstructed voxel representation of the scene. Deep learning-based approaches include \cite{gupta2014learning, wang2018depth, qi20173d}. 
\cite{gupta2014learning} proposed the HHA encoding of depth images. The authors used this encoding to design an object detection system, which they further exploited to improve semantic segmentation performance.
\cite{shelhamer2017fully} also used the HHA in their pioneering work on FCNs.
\cite{wang2018depth} proposed a depth-aware convolution and pooling mechanism to incorporate geometry into the convolution operators to build a depth-aware receptive field. 
\cite{qi20173d} used the output of a 2D segmentation network to initialize node features of a graph neural network applied on a 3D point cloud which was backprojected from a depth image. In this work, we also leverage RGB-D images, but focus on segmenting each individual object with unknown object class.

\begin{figure*}[t]
\begin{center}
\includegraphics[width=\linewidth]{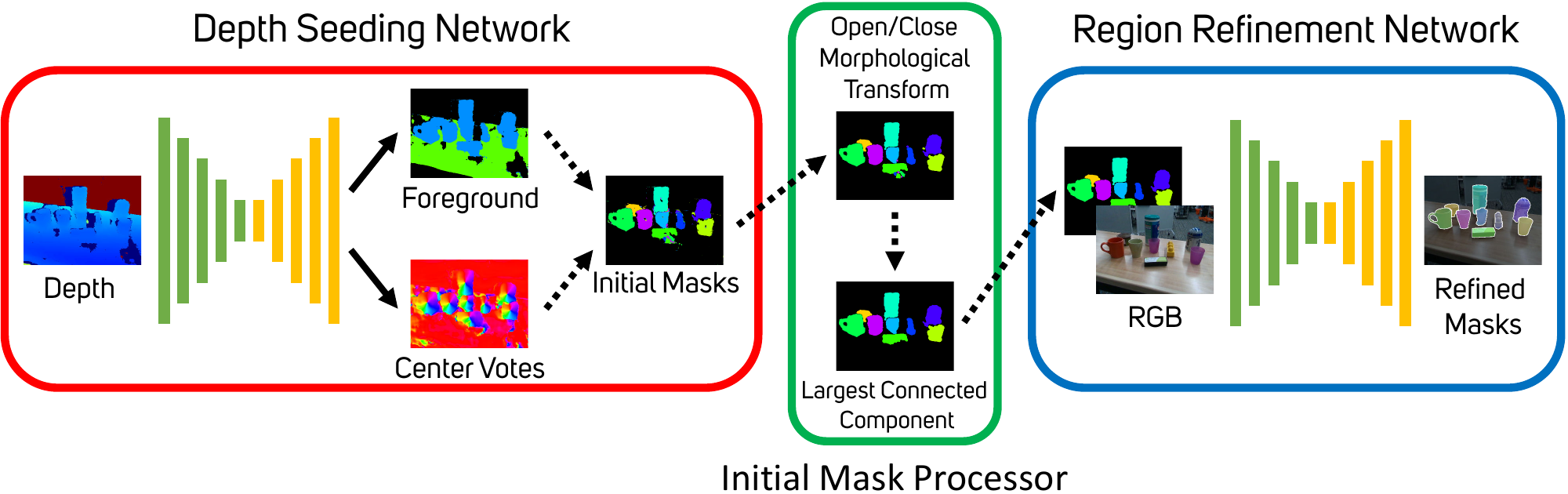}
\caption{Overall architecture. The Depth Seeding Network (\textcolor{red}{DSN}) is shown in the red box, the Initial Mask Processor (\textcolor{green}{IMP}) in the green box, and the Region Refinement Network (\textcolor{blue}{RRN}) in the blue box. The images come from a real example taken by an RGB-D camera in our lab. Despite the level of noise in the depth image (due to reflective table surface), our method is able to produce sharp and accurate instance masks. Gradients do not flow backwards through dotted lines.}
\label{fig:architecture}
\end{center}
\end{figure*}

\subsection{Instance-level Object Segmentation}

2D object instance segmentation is the problem of segmenting every object instance in an image. Many approaches for this problem involve top-down solutions that combine segmentation with object proposals in the form of bounding boxes \cite{he2017mask, li2017fully, Chen_2018_CVPR, kirillov2020pointrend}, typically produced by a region proposal network (RPN). FCIS \cite{li2017fully} utilizes position-sensitive inside/outside score maps for fully end-to-end convolutional instance segmentation. Mask R-CNN \cite{he2017mask}, a prominent work in the field, predicts a foreground mask for each object proposal. \cite{Chen_2018_CVPR} builds on top of both FCIS and Mask R-CNN and exploits semantic segmentation and direction predictions to assemble foreground masks. \cite{kirillov2020pointrend} proposes a module that iteratively refines
segmentation predictions at adaptively selected locations, which can be used in conjunction with Mask R-CNN.

However, when bounding boxes contain multiple objects (e.g. cluttered robot manipulation setups), the true instance mask is ambiguous and these methods struggle. Recently, a few methods have investigated bottom-up methods which assign pixels to object instances \cite{de2017semantic, Neven_2019_CVPR, Novotny_2018_ECCV, shao2018clusternet}. Other methods examine dense sliding-window instance segmentation on 4D tensors \cite{chen2019tensormask}, combining top-down and bottom-up methods via blending modules \cite{chen2020blendmask}, and alternative mask representations such as contours \cite{peng2020deep}. Additionally, interactive instance segmentation has shown strong results with few user inputs \cite{maninis2018deep}.


Most of the afore-mentioned algorithms provide instance masks with category-level semantic labels, which do not generalize to unseen objects in novel categories.
One approach to adapting these techniques to unseen objects is to employ ``class-agnostic'' training, which treats all object classes as one foreground category \cite{danielczuk2018segmenting}. One family of methods exploits motion cues with class-agnostic training in order to segment arbitrary moving objects \cite{xie2019object, dave2019towards}. Another family of methods are class-agnostic object proposal algorithms \cite{DeepMask, SharpMask, kuo2019shapemask}. However, these methods will segment everything and require some post-processing method to select the masks of interest. \cite{shao2018motion} jointly estimates instance segmentation masks and rigid scene flow, similar to \cite{byravan2017se3nets,byravan2018se3posenets}. We also train our proposed method in a class-agnostic fashion, but instead focus our notion of unseen objects in particular environments such as tabletop settings.

In 3D instance segmentation, researchers have recently been investigating architectures to apply on point clouds/voxel grids. \cite{hou20193d} introduced the first deep learning method to fuse RGB and geometric information from RGB-D scans. \cite{han2020occuseg} proposed an occupancy term which greatly aids supervoxel clustering, leading to strong results. A few of these methods embrace center voting-based techniques \cite{lahoud20193d, qi2019deep, qi2020imvotenet, engelmann20203d, jiang2020pointgroup}. \cite{lahoud20193d} utilizes metric learning to learn abstract features, and predicts center votes which are post-processed by meanshift clustering. \cite{qi2019deep} uses the center votes with a simple grouping mechanism to detect 3D bounding boxes of objects from point clouds only. Their follow up work \cite{qi2020imvotenet} incorporates RGB information by lifting 2D votes and features into 3D. \cite{engelmann20203d} follows a similar architecture but stacks a graph convolutional network to refine proposal features. \cite{jiang2020pointgroup} also performs clustering on votes and semantic features for targeted performance on certain object classes. Our method takes inspiration from these voting-based methods, but is targeted to cluttered robot environments.


\subsection{Sim-to-Real Perception}

Training a model on synthetic RGB and directly applying it to real data typically fails \cite{zhang2015towards}. Many methods employ some level of rendering randomization \cite{xiang2017posecnn, tremblay2019deep, li2017deepim, tobin2017domain, pinto2017asymmetric, sadeghi2017cadrl}, including lighting conditions and textures. However, they typically assume specific object instances and/or known object models. Another family of methods employ domain adaptation to bridge the gap between simulated and real images \cite{tzeng2015adapting, bousmalis2018using}. Algorithms trained on depth have been shown to generalize reasonably well for simple settings \cite{mahler2017dex, danielczuk2018segmenting}. However, noisy depth sensors can limit the application of such methods. Our proposed method is trained purely on (non-photorealistic) synthetic RGB-D data and is accurate even when depth sensors are inaccurate, and can be trained without adapting or randomizing the synthetic RGB.

%% file: method_overview.tex
\section{Method Overview}

Given a single RGB-D image, the goal of our algorithm is to produce object instance segmentation masks for all objects on a tabletop, where the object instances (or even the semantic class) are arbitrary and are not assumed to have been seen during a training phase. These masks do not have any notion of class categorization or semantics. These masks can be employed by robots for interacting with unseen object instances in downstream applications such as grasping and/or manipulation. 

Our framework consists of two separate networks that process Depth and RGB separately to produce instance segmentation masks. First, we design a Depth Seeding Network (DSN) that takes a depth image as input and outputs \textit{initial} object instance segmentation masks. These initial masks can be quite noisy for a number of reasons, thus we design an Initial Mask Processor (IMP) to robustify them with standard image processing techniques. We further refine the processed initial masks using our Region Refinement Network (RRN), which is designed to snap the noisy initial mask edges to object edges in RGB, providing sharp and accurate final instance masks. The full architecture is shown in Figure \ref{fig:architecture}.

Because the DSN incorporates non-differentiable techniques in order to build the initial masks, our DSN and RRN are trained separately as opposed to end-to-end. Both the DSN and RRN can be trained fully in simulation with no fine-tuning on real-world data, allowing our framework to capitalize on large amounts of simulated scenes and objects without resorting to the expensive process of annotating data. Our framework generalizes remarkably well to real-world scenarios despite being trained only on non-photorealistic simulated data, enabling robotic tasks with unseen objects.

%% file: dsn.tex
\section{Depth Seeding Network}

It has been shown that depth generalizes reasonably well for Sim-to-Real problems \cite{mahler2017dex, danielczuk2018segmenting, seita2018robot}. Inspired by this concept, we focus the first stage of our framework on depth only to produce initial class-agnostic instance segmentation masks. At a high level, the DSN takes as input a 3-channel organized point cloud, $D \in \R^{H \times W \times 3}$, of XYZ coordinates, and outputs initial instance segmentation masks. Note that $D$ can be computed by backprojecting a depth map given camera intrinsics.

We examine two methods of structuring the DSN. First, we investigate building initial masks by predicting centers in 2D pixel space. While this method provides state-of-the-art results, it has some obvious pitfalls (examined in Section \ref{sec:experiments}) that motivates a novel architecture that builds masks by predicting centers in 3D space. 

\subsection{Reasoning in 2D}

\subsubsection{Network Architecture}

The organized point cloud $D$ is passed through an encoder-decoder architecture to produce two outputs: a semantic segmentation mask $F \in \R^{H \times W \times C}$, where $C$ is the number of semantic classes, and 2D directions to object centers $V \in \R^{H \times W \times 2}$. We use $C=3$ for our semantic classes: background, tabletop, and tabletop objects. Each pixel of $V$ encodes a 2-dimensional unit vector pointing to the 2D center of the object. We define the center of the object to be the mean pixel location of the observable mask (part of mask that is unoccluded). Although we do not explicitly make use of the tabletop label in Section \ref{sec:experiments}, it can be used in conjunction with RANSAC \cite{fischler1981random} in order to better estimate the table for downstream applications. For the encoder-decoder architecture, we use a U-Net \cite{ronneberger2015u} architecture where each $3 \times 3$ convolutional layer is followed by a GroupNorm layer \cite{wu2018group} and ReLU. The output of the U-Net is a feature map of shape $\R^{H \times W \times 64}$. Sitting on top of this is two parallel branches of convolutional layers that produce the foreground mask $F$ and center directions $V$ (Figure~\ref{fig:architecture}).

\removed{While we use U-Net for the DSN architecture, our framework is not limited to this and can replace it with any network architecture.}

In order to compute the initial segmentation masks from $F$ and $V$, we design a Hough voting layer similar to \cite{xiang2017posecnn}. We describe the pseudocode detailed in Algorithm~\ref{alg:hough_voting}. First, we discretize the space of angles $[0, 2\pi]$ into $A$ equally spaced bins. For every pixel, we compute the percentage of discretized directions from all other foreground pixels that point to it and use this as a score for how likely the pixel is an object center (lines 3-10). We threshold when a foreground pixel points to it with an inlier threshold and distance threshold. We then threshold the percentages and apply non-maximum suppression (NMS) to select object centers (line 11). Given these object centers, each pixel is assigned to the closest center it points to (line 12), which gives the initial masks as shown in the red box of Figure~\ref{fig:architecture}. Note that the inlier, distance, and percentage thresholds provide the Hough voting layer with robustness. For example, if not enough foreground pixels from all directions point towards a potential object center, that center is not selected. This robustifies the algorithm by protecting against false positives. We qualitatively show the efficacy of these design choices in Section~\ref{subsec:qualitative_2D}.

\subsubsection{Loss Functions}
\label{subsubsec:2D_DSN_loss_function}

To train the DSN, we apply two different loss functions on the semantic segmentation $F$ and the direction prediction $V$. 

\paragraph{Foreground Loss} For the semantic segmentation $F$, we use a weighted cross entropy as this has been shown to work well in detecting object boundaries in imbalanced images \cite{xie2015holistically}. The loss is 
$
\ell_{fg} = \sum_i w_i\ \ell_{ce}\left(F_i, \bar{F}_i\right)$
where $i$ ranges over pixels, $F_i, \bar{F}_i$ are the predicted and ground truth probabilities of pixel $i$, respectively, and $\ell_{ce}$ is the cross-entropy loss. The weight $w_i$ is inversely proportional to the number of pixels with labels equal to $\bar{F}_i$, normalized to sum to 1.

\paragraph{Direction Loss} We apply a weighted cosine similarity loss to the direction prediction $V$. The cosine similarity is focused on the tabletop object pixels, but we also apply it to the background/tabletop pixels to have them point in a fixed direction to avoid false positives. The loss is given by 

\begin{equation}
\ell_{dir} = \sum_{i \in \mathcal{O}} \alpha_i \left(1 - V_i^\T \bar{V}_i\right) + \frac{\lambda_{bt}}{|\mathcal{B} \cup \mathcal{T}|} \sum_{i \in B \cup T} \left(1 - V_i^\T \colvec{0 \\ 1} \right) \,,
\end{equation}
where $V_i, \bar{V}_i$ are the predicted and ground truth unit directions of pixel $i$, respectively. $\mathcal{B}, \mathcal{T}, \mathcal{O}$ are the sets of pixels belonging to background, table, and object/foreground classes, respectively. Note that $\mathcal{B} \cup \mathcal{T} \cup \mathcal{O} = \Omega$, where $\Omega$ is the set of all pixels. $\alpha_i$ is inversely proportional to the number of pixels with the same \textit{instance} label as pixel $i$, which gives equal weight to each instance regardless of size. We set $\lambda_{bt} = 0.1$. The total loss for our 2D DSN is given by $\ell_{fg} + \ell_{dir}$.

\begin{algorithm}[!t]
\caption{Hough Voting Procedure Pseudocode}
\label{alg:hough_voting}
\begin{algorithmic}[1]
\REQUIRE $F$, $V$, cosine distance $d_c$, number of angle bins $A$. Robustness parameters: inlier threshold $\eps_{it}$, distance threshold $\eps_{d}$, percentage threshold $\eps_{pt}$.
\RETURN Initial instance masks $S$

\STATE Initialize $\mathcal{H} \in \R^{H \times W \times A}$ to zeros
\FOR{potential center $p_c \in \Omega$}
\FOR{$p \in F$}
\IF{$d_c(p_c-p,V_{p}) < \eps_{it}$ AND $d(p_c,p) < \eps_{d}$}
\STATE $a = \left\lfloor A \cdot d_c\left(p_c-p, [0,1]\right) \right\rfloor$ \# discretized angle
\STATE $\mathcal{H}_{p_c, a} = 1$
\ENDIF
\ENDFOR
\ENDFOR
\STATE Compute local maximums $\mathcal{C}$ of $\mathcal{H}$ using NMS and $\eps_{pt}$
\STATE Compute $S$ with $\mathcal{C},V$
\end{algorithmic}
\end{algorithm}

\subsection{Reasoning in 3D}

Reasoning in 2D has some failure cases that can be mitigated by reasoning in 3D. For example, if the center of an object is occluded by another object, the 2D center voting procedure will not detect that object (examples of this can be found in Section~\ref{subsec:qualitative_2D}). Thus, we propose a new architecture to the DSN to better handle these cases and provide stronger results. This formulation requires more sophisticated loss functions. In particular, we introduce a novel separation loss that significantly improves accuracy in cluttered scenes.

\subsubsection{Network Architecture}

The network architecture of this 3D DSN is almost the same as the 2D DSN. The input is the same, which is the organized point cloud $D$ of XYZ coordinates. The main difference between the 2D DSN and the 3D DSN is the output. Our 3D DSN still outputs the semantic segmentation mask $F$,
but produces 3D offsets to object centers $V' \in \R^{H \times W \times 3}$ instead of 2D directions $V$. Note that elements in $V'$ are not unit vectors, which is the case with $V$. Since $V'$ are 3D offsets, $D+V'$ is the predicted object centers for each pixel, which we will refer to as ``center votes''. 

We propose to modify the architecture to use dilated convolutions~\cite{Yu2016multiscale} in order to provide the DSN with a higher receptive field. We replace the $6^\textrm{th}, 8^\textrm{th}$, and $10^\textrm{th}$ convolution layers with ESP modules~\cite{mehta2018espnet}. An ESP module is a lightweight module consisting of a \textit{reduction} operation, a \textit{split/transform} component that applies convolutions with different dilation rates to get a spatial pyramid, and a \textit{merge} process that hierarchically fuses the feature maps of the spatial pyramid~\cite{mehta2018espnet}. The ESP module has less parameters than the convolution layer it replaces, making it more computationally efficient. Details of the implementation can be found in the public code release at the project website\footnote{\url{https://rse-lab.cs.washington.edu/projects/unseen-object-instance-segmentation/}}. We show in Section~\ref{subsec:quantitative_3D} that adding this module provides a boost in performance.

To compute initial masks, we perform mean shift clustering in 3D space over our center votes $D+V'$. Mean shift clustering is an iterative procedure to find the modes of a distribution approximated by a kernel density esimate (KDE). The number of clusters (objects, in our case) is not determined beforehand, but instead by the number of modes in the KDE. We use the Gaussian kernel $K(x,y) = \exp\left( \frac{1}{\sigma^2} \|x-y\|_2^2 \right)$, which results in Gaussian mean shift (GMS) clustering. $\sigma > 0$ is a hyperparameter which affects the number of modes (objects) in the KDE. Thus, the choice of $\sigma$ is crucial and depends on the relative distance between objects, which is low in clutter. A detailed review of mean shift clustering algorithms can be found in~\cite{carreira2015review}. After clustering, each pixel is assigned to the cluster ID of its center vote to generate the initial masks. The clustering is only applied to the foreground pixels. Note that this method of producing initial masks lacks the thresholds such as $\eps_{it}, \eps_d, \eps_{pt}$ from the 2D DSN Hough voting layer that provide robustness. 

\subsubsection{Loss Functions}

We apply four loss functions on the semantic segmentation $F$ and center offsets $V'$ to train the 3D-reasoning version of the Depth Seeding Network. 

\paragraph{Foreground Loss} 

We utilize the same foreground loss as in Section~\ref{subsubsec:2D_DSN_loss_function} for the 2D DSN, $\ell_{fg}$.

\paragraph{Center Offset Loss} 

We apply a Huber loss $\rho$ (Smooth L1 loss) to the center offsets $V'$ to penalize the distance of the center votes to their corresponding ground truth object centers. 
\begin{equation}
    \ell_{co} = \sum_{i \in \Omega} w_i \rho\left( D_i + V'_i - c_i \right) \,,
\end{equation}
where $c_i$ is the 3D coordinate of the ground truth object center for pixel $i$. Like $\ell_{fg}$, the weight $w_i$ is inversely proportional to the number of pixels with the same instance label $y_i$. For object centers that are out of view of the camera, we project them to the camera's field of view.

\begin{figure*}[t]
\begin{center}
\includegraphics[width=\linewidth]{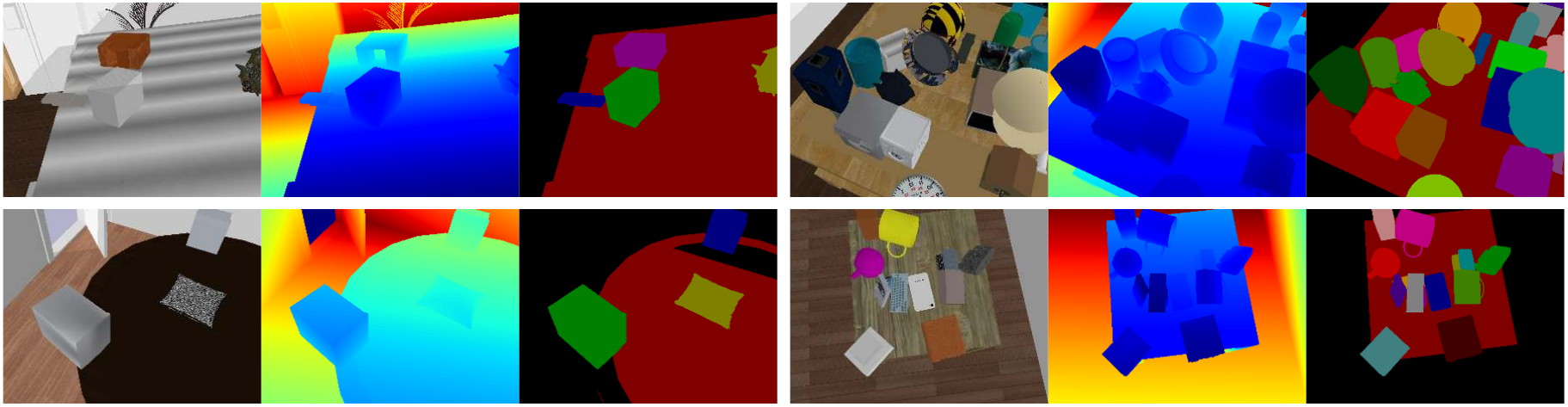}
\caption{Examples from our Tabletop Object Dataset. (Non-photorealistic) RGB, depth, and instance masks are shown.}
\label{fig:tod_image}
\end{center}
\end{figure*}

\paragraph{Clustering Loss}

We adopt a clustering loss that unrolls GMS for a few iterations and applies a loss on the clustered points, very similar to~\cite{kong2018recurrent}.
GMS iteratively shifts a set of $S$ 3D seed points, $Z \in \R^{S \times 3}$, to higher density regions of the KDE in a gradient ascent-type fashion~\cite{carreira2015review}. Let $Z^{(l)}$ be the points at the $l^\textrm{th}$ iteration of GMS. $Z^{(0)}$ is initialized as the center votes $X = D + V' \in \R^{|\mathcal{O}| \times 3}$ of the foreground pixels. One iteration of GMS amounts to 
$Z^{(l+1)} = \tilde{D}^{-1} K X$
where $K \in \R^{S \times |\mathcal{O}|}$ is the kernel matrix s.t. $[K]_{ij} = K(Z^{(l)}_i,X_j)$, and $\tilde{D} = \textrm{diag}(K \mathbf{1})$ with $\mathbf{1} \in \R^{|\mathcal{O}|}$ being a vector of all ones. Note that $K$ depends on $\sigma$. This iteration can be seen as a layer in the network with no parameters for learning. We apply the following loss function to $Z^{(l)}$ and $X$, with the corresponding object instance labels $Y \in \R^{|\mathcal{O}|}$:
\begin{align}
\label{eq:cluster_loss_full}
\begin{split}
    \ell_{cl}^{(l)} (Z^{(l)},X,Y) = \sum_{i=1}^S \sum_{j \in \mathcal{O}} &w_{ij} \mathbbm{1}\{y_i = y_j\} d^2(Z^{(l)}_i,X_j) \\
    + w_{ij} &\mathbbm{1}\{y_i \neq y_j\} [\delta - d(Z^{(l)}_i,X_j)]_+^2
\end{split}
\end{align}
where $w_{ij}$ are inverse proportional weights w.r.t. class size, $d(\cdot,\cdot)$ is Euclidean distance, and $[\cdot]_+ = \textrm{max}(\cdot, 0)$.
This loss function influences the KDE modes to be close to its points, and at least $\delta$ away from points not belonging to the cluster, encouraging the points $X = D+V'$ to be more cluster-like. 

Applying Eq. (\ref{eq:cluster_loss_full}) to all points, i.e. $S = |\mathcal{O}|$, results in excessive memory usage, thus we instead adopt a stochastic version of this loss function. We randomly sample an index set $\mathcal{I} \subset \{1, 2, \ldots, |\mathcal{O}|\}$ and set $Z^{(0)} = X_{\mathcal{I}}$, and run GMS clustering only on these points. We unroll GMS for $L$ iterations, and apply $\ell_{cl}^{(l)}$ at each iteration, giving the full cluster loss $\ell_{cl} = \ell_{cl}^{(1)} + \ldots + \ell_{cl}^{(L)}$.

\paragraph{Separation loss}

We introduce a novel separation loss that encourages the center votes to not necessarily be at the center of an object, as long as it is far away from other object center votes \revisiontext{in order to ease the post-processing GMS clustering phase}. 
To do this, we consider the following tensor:
\begin{equation}\label{eq:sep_loss}
    M_{ij} = \frac{\exp\left(-\tau\ d(c_j, D_i+V'_i)\right)}{\sum_{j'=1}^J \exp\left(-\tau\ d(c_{j'}, D_i+V'_i)\right)}
\end{equation}
where $c_j$ is the $j^\mathrm{th}$ ground truth object center, $i \in \mathcal{O}$, and $\tau > 0$ is a hyperparameter. \revisiontext{This is simply the distance from center vote $D_i + V'_i$ to all object centers scaled by $\tau$, with a softmax applied. We apply a cross entropy loss $\ell_{sep}(M_{ij}) = - \sum_{j=1}^N \mathbbm{1}\{y_i = j\} \log(M_{ij})$ in order to maximize $M_{ij}$ when $y_i = j$.}

Maximizing $M_{ij}$ for a foreground pixel $i$ and its corresponding GT object center $j$ encourages 1) the center vote $D_i+V'_i$ to be close to $c_j$, and 2) to be far from $\{c_j\ |\ y_i \neq y_j\}$. While the $\ell_{co}$ also enforces property 1, property 2 is quite desirable in heavy clutter. If two objects are situated in a way such that their 3D object centers are very close, post-processing clustering will be difficult. This separation loss encourages the network to predict object center votes that are not close to each other, making the task of post-processing clustering easier. In Section~\ref{subsec:quantitative_3D}, we show that this loss is crucial for strong performance in heavy clutter.

In summary, the total loss used to train the 3D DSN is given by $\lambda_{fg} \ell_{fg} + \lambda_{co} \ell_{co} + \lambda_{cl} \ell_{cl} + \lambda_{sep} \ell_{sep}$. 

%% file: imp.tex
\section{Initial Mask Processing Module}

Computing the initial masks from $F$ and $V/V'$ often results in noisy masks (see an example of initial masks computed from the 2D DSN using our Hough voting layer in Figure \ref{fig:architecture}). For example, these instance masks often exhibit salt/pepper noise and erroneous holes near the object center (see Section \ref{subsec:qualitative_2D} for examples). As shown in Section \ref{subsec:quantitative_2D}, the RRN has trouble refining the masks when they are scattered as such. To robustify the algorithm, we propose to use two simple image processing techniques to clean the masks before refinement.

For a single instance mask, we first apply an opening operation, which consists of mask erosion followed by mask dilation \cite{serra1983image}, removing the salt/pepper noise issues. Next we apply a closing operation, which is dilation followed by erosion, which closes up small holes in the mask. Finally, we select the largest connected component and discard all other components. Note that these operations are applied to each instance mask separately. These simple image processing techniques are immensely helpful in robustifying the system.

%% file: rrn.tex
\section{Region Refinement Network}

While depth generalizes reasonably well from Sim-to-Real, the initial masks (after IMP) are still subject to many errors due to noisy depth sensors. The RRN is designed to \textit{snap} the initial mask edges to the object edges in RGB to provide accurate and sharp instance masks.

\subsection{Network Architecture}

Inspired by \cite{maninis2018deep}, this network takes as input a cropped 4-channel image, which consists of RGB concatenated with a single initial instance mask. The RGB image is cropped around the instance mask with some padding for context, concatenated with the (cropped) mask, then resized to $224 \times 224$. This gives an input image $I \in \R^{224 \times 224 \times 4}$. The output of the RRN is the refined mask probabilities $R \in \R^{224 \times 224}$, which we threshold to get the final output. We use the same U-Net architecture as in the DSN. To train the RRN, we apply the loss $\ell_{fg}$ with two classes (foreground vs. background) instead of three.

\subsection{Mask Augmentation}
\label{subsubsec:mask_augmentation}

Recall that the DSN and RRN are trained separately. In order to train the RRN, we need examples of perturbed instance masks. \revisiontext{While we could train the RRN with the outputs of the DSN, we found that they are typically too clean on our synthetic dataset and we achieved better results by perturbing the ground truth masks instead.} This problem can be seen as a data augmentation task where we augment the mask into something that resembles an initial mask (after the IMP). We detail the different augmentation techniques used below:

\begin{itemize}
    
    \item Translation/rotation: We translate the mask by sampling a displacement vector proportionally to the mask size. \removed{from a beta distribution.} Rotation angles are sampled uniformly in $[-10\degree, 10\degree]$.
    
    \item Adding/cutting: For this augmentation, we choose a random part of the mask near the edge, and either remove it (cut) or copy it outside of the mask (add). This reflects the setting when the initial mask egregiously overflows from the object, or is only covering part of it.
    
    \item Morphological operations: We randomly choose multiple iterations of either erosion or dilation of the mask. The erosion/dilation kernel size is set to be a percentage of the mask size, where the percentage is sampled from a beta distribution. This reflects inaccurate boundaries in the initial mask, e.g. due to noisy depth sensors.
    
    \item Random ellipses: We sample the number of ellipses to add or remove in the mask from a Poisson distribution. For each ellipse, we sample both radii from a gamma distribution and a random rotation angle. This augmentation requires the RRN to learn to remove irrelevant blots outside of the object and close up small holes within it.
    
\end{itemize}

%% file: tod.tex
\section{Tabletop Object Dataset}

Many desired robot environment settings (e.g. kitchens, cabinets) lack large scale training data to train deep networks. To our knowledge, there is no large scale dataset for unseen tabletop objects. To remedy this, we generate our own synthetic dataset which we name the Tabletop Object Dataset (TOD). This dataset is comprised of 40k synthetic scenes of cluttered ShapeNet~\cite{shapenet2015} objects on a (ShapeNet) tabletop in SUNCG home environments \cite{song2016ssc}. 
We only use ShapeNet tables that have convex tabletops and filter the ShapeNet object classes to roughly 25 classes of objects that could potentially be on a table. Example classes include: jar, mug, helmet, and pillow.

Each scene in the dataset is of a random room chosen from a random SUNCG house loaded without any furniture. We sample a ShapeNet table and scale it such that its height is in the range [0.75m, 1m], and place it in the room so that it is not colliding with walls or other fixtures in the room. Next, we randomly sample anywhere between 5 and 25 objects to put on the table, rescaling them such that they are not larger than $\frac14\min\left\{t_h,t_l\right\}$ where $t_h,t_l$ are the height and length of the table, respectively. The objects are either randomly placed on the table, on top of another object (stacked), or generated at a random height and orientation above the table. We use PyBullet \cite{coumans2019} to simulate physics until the objects come to rest and remove any objects that fell off the table. Next, we generate seven views: one view is of only background, another is of just the table in the room, and the rest are taken with random camera viewpoints with the tabletop objects in view. The viewpoints are sampled at a height of between .5m and 1.2m above the table and rotated randomly with an angle in $[-12\degree, 12\degree]$. The images are generated at a resolution of $640 \times 480$ with vertical field-of-view of 45 degrees. The segmentation has a tabletop (table plane only) label and instance labels for each object.

We show some example images of our dataset in Figure \ref{fig:tod_image}. The rightmost two examples show that some of our scenes are heavily cluttered. Note that the RGB looks non-photorealistic. In particular, PyBullet is unable to load textures of some ShapeNet objects (see gray objects in leftmost two images). PyBullet was built for reinforcement learning, not computer vision, thus its rendering capabilities are insufficient for photorealistic tasks \cite{coumans2019}. Despite this, our RRN can learn to snap masks to object boundaries from this synthetic dataset.

\postsec

%% file: experiments.tex
\section{Experiments}

\label{sec:experiments}
\presec

\begin{table}[t]
\centering
\begin{tabular}{|c|ccc|ccc|}
\hline
\multirow{2}{*}{Method} & \multicolumn{3}{c|}{Overlap} & \multicolumn{3}{c|}{Boundary} \\
 & \textcolor{orange}{P} & \textcolor{cyan}{R} & \textcolor{purple}{F} & \textcolor{orange}{P} & \textcolor{cyan}{R} & \textcolor{purple}{F}\\
\hline
GCUT \cite{felzenszwalb2004efficient} & 21.5 & 51.5 & {25.7} & 10.2 & 46.8 & {15.7} \\
SCUT \cite{pham2018scenecut} & 45.7 & 72.5 & {43.7} & 43.1 & 65.1 & {42.6} \\
LCCP \cite{christoph2014object} & 58.4 & \textcolor{red}{89.1} & {63.8} & 53.6 & \textcolor{red}{82.6} & {60.2} \\
V4R \cite{potapova2014incremental} & 65.3 & 81.4 & {69.5} & 62.5 & 81.4 & {66.6} \\
UOIS-Net-2D & \textcolor{red}{88.8} & 81.7 & 84.1 & \textcolor{red}{83.0} & 67.2 & 73.3 \\
UOIS-Net-3D & 88.2 & 88.0 & {\textcolor{red}{87.9}} & 81.1 & 74.3 & {\textcolor{red}{77.3}} \\
\hline
\end{tabular}
\caption{Comparison with baselines on ARID20 and YCB10 subsets of OCID \cite{suchi2019easylabel}. \textcolor{red}{Red} indicates the best performance.}
\label{table:baseline_comparison}
\end{table}

We evaluate our method on real datasets against the state-of-the-art (SOTA) methods Mask R-CNN~\cite{he2017mask} \revisiontext{and PointGroup~\cite{jiang2020pointgroup}}. We denote our method as \textit{UOIS-Net-2D} or \textit{UOIS-Net-3D}, depending on whether the DSN reasons in either 2D or 3D. 

\begin{table*}[t]
\centering

\begin{tabular}{|c|c|ccc|ccc|ccc|ccc|}
\hline
\multirow{3}{*}{Method} & \multirow{3}{*}{Input} & \multicolumn{6}{c|}{OCID \cite{suchi2019easylabel}} & \multicolumn{6}{c|}{OSD \cite{richtsfeld2012segmentation}} \\ \cline{3-14}
 &  & \multicolumn{3}{c|}{Overlap} & \multicolumn{3}{c|}{Boundary} & \multicolumn{3}{c|}{Overlap} & \multicolumn{3}{c|}{Boundary} \\
 &  & \textcolor{orange}{P} & \textcolor{cyan}{R} & \textcolor{purple}{F} & \textcolor{orange}{P} & \textcolor{cyan}{R} & \textcolor{purple}{F} & \textcolor{orange}{P} & \textcolor{cyan}{R} & \textcolor{purple}{F} & \textcolor{orange}{P} & \textcolor{cyan}{R} & \textcolor{purple}{F} \\ \hline
Mask R-CNN~\cite{he2017mask} & RGB & 66.0 & 34.0 & 36.6 & 58.2 & 25.8 & 29.0 & 63.7 & 43.1 & 46.0 & 46.7 & 26.3 & 29.5 \\
Mask R-CNN~\cite{he2017mask} & Depth & 82.7 & 78.9 & 79.9 & 79.4 & 67.7 & 71.9 & 73.8 & 72.9 & 72.2 & 49.6 & 40.3 & 43.1  \\
Mask R-CNN~\cite{he2017mask} & RGB-D & 79.2 & 78.6 & 78.0 & 73.6 & 67.2 & 69.2 & 74.0 & 74.6 & 74.1 & 57.3 & 52.1 & 53.8 \\
\revisiontext{PointGroup~\cite{jiang2020pointgroup}} & \revisiontext{RGB-D} & \revisiontext{81.6} & \revisiontext{80.1} & \revisiontext{80.1} & \revisiontext{75.4} & \revisiontext{70.4} & \revisiontext{71.7} & \revisiontext{79.5} & \revisiontext{78.2} & \revisiontext{78.8} & \revisiontext{67.0} & \revisiontext{65.0} & \revisiontext{65.4} \\
UOIS-Net-2D & DSN: RGB-D & 84.2 & 57.6 & 62.2 & 72.9 & 44.5 & 49.9 & 72.2 & 63.7 & 66.1 & 58.5 & 43.4 & 48.4 \\
UOIS-Net-2D & DSN: Depth & \textcolor{red}{88.3} & 78.9 & 81.7 & \textcolor{red}{82.0} & 65.9 & 71.4 & 80.7 & 80.5 & 79.9 & 66.0 & 67.1 & 65.6 \\
UOIS-Net-3D & DSN: Depth & 86.5 & \textcolor{red}{86.6} & \textcolor{red}{86.4} & 80.0 & \textcolor{red}{73.4} & \textcolor{red}{76.2} & \textcolor{red}{85.7} & \textcolor{red}{82.5} & \textcolor{red}{83.3} & \textcolor{red}{75.7} & \textcolor{red}{68.9} & \textcolor{red}{71.2} \\
\hline

\end{tabular}
\caption{Evaluation of our methods against SOTA methods trained on different input modes. \textcolor{red}{Red} indicates the best performance.}
\label{table:input_comparison_sota}
\end{table*}

\begin{table*}[t]
\centering
\begin{tabular}{|c|ccc|ccc|ccc|ccc|}
\hline
\multirow{3}{*}{Method} & \multicolumn{6}{c|}{OCID~\cite{suchi2019easylabel}} & \multicolumn{6}{c|}{OSD~\cite{richtsfeld2012segmentation}} \\ \cline{2-13}
& \multicolumn{3}{c|}{Overlap} & \multicolumn{3}{c|}{Boundary} & \multicolumn{3}{c|}{Overlap} & \multicolumn{3}{c|}{Boundary} \\
& \textcolor{orange}{P} & \textcolor{cyan}{R} & \textcolor{purple}{F} & \textcolor{orange}{P} & \textcolor{cyan}{R} & \textcolor{purple}{F} & \textcolor{orange}{P} & \textcolor{cyan}{R} & \textcolor{purple}{F} & \textcolor{orange}{P} & \textcolor{cyan}{R} & \textcolor{purple}{F} \\ \hline
UOIS-Net-2D & \revisiontext{90.9} & \revisiontext{79.4} & \revisiontext{83.0} & 85.2 & 70.8 & 75.7 & \revisiontext{80.3} & \revisiontext{81.1} & \revisiontext{80.1} & 53.4 & 53.3 & 52.8 \\
UOIS-Net-3D & \revisiontext{88.8} & \revisiontext{89.2} & \revisiontext{88.8} & 86.9 & 80.9 & 83.5 & \revisiontext{85.7} & \revisiontext{82.1} & \revisiontext{83.0} & 74.3 & 67.5 & 70.0 \\
\hline
\end{tabular}
\caption{Performance of UOIS-Net without RRN.}
\label{table:OCID_issue}
\end{table*}

\subsection{Implementation Details}

All images have resolution $H = 480, W = 640$. 

\subsubsection{DSN}

We train all 2D DSN models for 100k iterations with stochastic gradient descent (SGD) using a fixed learning rate of 1e-2. We use a batch size of 8 and set \revisiontext{$\lambda_{fg} = \lambda_{dir} = 1$}. For the Hough voting layer, we use discretize the angles into $A=100$ bins, and set $\eps_{it} = 0.9, \eps_{d} = 20, \eps_{pt} = 0.5$. We also process every 10th pixel instead of every pixel in line 4 of Algorithm~\ref{alg:hough_voting} for computational efficiency. 

All 3D DSNs are trained for 150k iterations with Adam~\cite{kingma2015adam} with initial learning rate of 1e-4. We use a batch size of 8, with $\lambda_{fg}=3, \lambda_{co}=5, \lambda_{cl}=\lambda_{sep}=1$. In training, we rollout the cluster loss $\ell_{cl}$ $L = 5$ times and use $|\mathcal{I}| = 150$ seeds, while we use $L=10,|\mathcal{I}|=200$ during test time. $\delta$ is set to 0.1. We remove any cluster of pixels that is smaller than 500. Unless stated otherwise, we use $\tau = 15, \sigma = 0.02$. Note that we tried learning $\sigma$ as an output of the network, however it didn't perform well, thus we opted to keep it fixed.

During DSN training, we augment depth maps with multiplicative gamma noise similar to \cite{mahler2017dex}, and add Gaussian Process noise to the backprojected point clouds.

\subsubsection{RRN}

All RRNs are trained with SGD for 100k iterations with a fixed learning rate of 1e-2 and batch size of 16. 
\revisiontext{Inputs to the RRN are padded by 25\% of the initial mask's bounding box size in each dimension.}
\revisiontext{Our entire pipeline runs at approximately 3-5 frames per second on a single NVIDIA RTX 2080ti.}

\subsubsection{Baselines}

For the baselines in~\cite{suchi2019easylabel}, we use the results graciously provided by the authors. 
We follow the official Detectron schedule when training Mask R-CNN~\cite{he2017mask} on TOD, \revisiontext{and train for 100k iterations using SGD with a batch size of 8. We train PointGroup~\cite{jiang2020pointgroup} for 300k iterations with a batch size of 4 on TOD. We remove clustering of the semantic scores since our problem only has one meaningful semantic class, foreground, which makes no sense to cluster.}

Note that the training procedure for 2D DSNs and Mask R-CNN (reported in \cite{xie2019uois}) and differs slightly from the training procedure for 3D DSNs. We replicated these experiments using the same conditions as 3D DSNs (150k iterations with Adam/SGD), but observed very similar results to~\cite{xie2019uois}. Thus, we report numbers from~\cite{xie2019uois}.

\subsection{Datasets}

We evaluate quantitatively and qualitatively on two real-world datasets: OCID \cite{suchi2019easylabel} and OSD \cite{richtsfeld2012segmentation}, which have 2346 images of semi-automatically constructed labels and 111 manually labeled images, respectively. OSD is a small dataset that was manually annotated, so the annotation quality is high. OCID, which is much larger, uses a semi-automatic process of annotating the labels. 
It incrementally builds up the scene by adding one object at a time, and computes labels by calculating difference in depth.
However, this process is subject to depth sensor noise, so while the majority of the instance label is accurate, the label boundaries are noisy.
Additionally, OCID contains images with objects on a tabletop, and images with objects on a floor. Despite our method being trained in synthetic tabletop settings, it generalizes to floor settings as well.

Lastly, we use the Google Open Images Dataset v5~\cite{OpenImagesSegmentation} (OID) in an experiment to test the Sim-to-Real gap of the RRN. OID contains approximately 9 million ``in-the-wild'' images with image-level annotations, bounding boxes, and segmentations. In particular, it contains 2.8 million segmentations for 350 object classes. We filtered the 350 object object classes down to 156 classes that could potentially be on a tabletop, resulting in roughly 220k instance masks on real RGB images. 

\subsection{Metrics}

We use the precision/recall/F-measure (P/R/F) metrics as defined in \cite{dave2019towards}. These metrics promote methods that segment the desired objects and penalize methods that provide false positives. Specifically, the F-measure is computed between all pairs of predicted and ground truth masks, which are matched via the Hungarian method. Given a matching, the final P/R/F is computed by
\begin{equation}
P = \frac{\sum_i \left|s_i \cap g(s_i) \right|}{\sum_i \left|s_i\right|},
R = \frac{\sum_i \left|s_i \cap g(s_i) \right|}{\sum_j \left|g_j\right|},
F = \frac{2PR}{P+R}, \nonumber
\end{equation}
where $s_i$ denotes the set of pixels belonging to predicted object $i$, $g(s_i)$ is the set of pixels of the matched ground truth object of $s_i$, and $g_j$ is the set of pixels for ground truth object $j$. We denote this as Overlap P/R/F. 
See \cite{dave2019towards} for more details.

Segmentations with sharp or fuzzy boundaries will obtain similar Overlap P/R/F scores, which will not reflect the efficacy of the RRN.
To remedy this, we introduce a Boundary P/R/F measure to complement the Overlap P/R/F. Using the same Hungarian matching used to compute Overlap P/R/F, we compute Boundary P/R/F by
\begin{equation}
P = \frac{\sum_i \left|s_i \cap D\left[g(s_i)\right] \right|}{\sum_i \left|s_i\right|}, 
R = \frac{\sum_i \left|D\left[s_i\right] \cap g(s_i) \right|}{\sum_j \left|g_j\right|}, \nonumber
\end{equation}
where we overload notation and denote $s_i, g_j$ to be the set of pixels belonging to the boundaries of predicted object $i$ and ground truth object $j$, respectively. $D[\cdot]$ denotes the dilation operation, which allows for some slack in the prediction. \revisiontext{Note that these metrics are sensitive to the amount of allowed slack. However, without this, these numbers can look deceivingly low as no slack requires exact pixel boundary matching which can lead to issues with either noisy manual annotations or noisy semi-automatic annotations \cite{suchi2019easylabel}. For the dilation operation, we use a circular kernel where the diameter of the kernel depends on the size of the image.} Roughly, this metric is a combination of the $\mathcal{F}$-measure in \cite{Perazzi2016} along with the Overlap P/R/F as defined in \cite{dave2019towards}. 

We report all P/R/F measures in the range $[0,100]$ (P/R/F $\times 100$).

\subsection{2D Quantitative Results}

\label{subsec:quantitative_2D}


\begin{table*}[t]
\centering

\begin{tabular}{|c|c|ccc|ccc|ccc|ccc|}
\hline
\multirow{3}{*}{\parbox{1.75cm}{\centering DSN reasoning}} & \multirow{3}{*}{\parbox{1.75cm}{\centering RRN training data}} & \multicolumn{6}{c|}{OCID \cite{suchi2019easylabel}} & \multicolumn{6}{c|}{OSD \cite{richtsfeld2012segmentation}} \\ 
\cline{3-14} & & \multicolumn{3}{c|}{Overlap} & \multicolumn{3}{c|}{Boundary} & \multicolumn{3}{c|}{Overlap} & \multicolumn{3}{c|}{Boundary} \\
& & \textcolor{orange}{P} & \textcolor{cyan}{R} & \textcolor{purple}{F} & \textcolor{orange}{P} & \textcolor{cyan}{R} & \textcolor{purple}{F} & \textcolor{orange}{P} & \textcolor{cyan}{R} & \textcolor{purple}{F} & \textcolor{orange}{P} & \textcolor{cyan}{R} & \textcolor{purple}{F} \\ \hline
\multirow{2}{*}{\parbox{1.75cm}{\centering 2D}} & TOD & 88.3 & 78.9 & 81.7 & 82.0 & 65.9 & 71.4 & 80.7 & 80.5 & 79.9 & 66.0 & 67.1 & 65.6 \\
& OID & 87.9 & 79.6 & 81.7 & 84.0 & 69.1 & 74.1 & 81.2 & 83.3 & 81.7 & 69.8 & 73.7 & 70.8 \\ \hline
\multirow{2}{*}{\parbox{1.75cm}{\centering 3D}} & TOD & 86.5 & 86.6 & 86.4 & 80.0 & 73.4 & 76.2 & 85.7 & 82.5 & 83.3 & 75.7 & 68.9 & 71.2 \\
& OID & 86.0 & 88.2 & 86.9 & 83.1 & 77.6 & 79.9 & 86.0 & 85.1 & 84.8 & 81.0 & 75.3 & 77.3 \\
\hline

\end{tabular}
\caption{Comparison of RRN when training on TOD and real images from Google OID \cite{OpenImagesSegmentation}.}
\label{table:RRN_real_data}
\end{table*}

\begin{table*}[t]
\centering

\begin{minipage}[t]{.47\linewidth}
\centering
\begin{tabular}{|cccc|ccc|}
\hline
\multirow{2}{*}{DSN} & \multicolumn{2}{|c|}{IMP}  & \multirow{2}{*}{RRN} & \multicolumn{3}{c|}{Boundary} \\
& \multicolumn{0}{|c}{O/C} & \multicolumn{0}{c|}{CCC} & & \textcolor{orange}{P} & \textcolor{cyan}{R} & \textcolor{purple}{F} \\ \hline

\cmark &  &  &  & 35.0 & 58.5 & 43.4 \\
\cmark &  &  & \cmark & 36.0 & 48.1 & 39.6 \\
\cmark & \cmark &  &  & 49.2 & 55.3 & 51.7 \\
\cmark & \cmark &  & \cmark & 59.0 & 64.1 & 60.7 \\
\cmark & \cmark & \cmark &  & 53.4 & 53.3 & 52.8\\
\cmark & \cmark & \cmark & \cmark & 66.0 & 67.1 & 65.6 \\

\hline
\end{tabular}
\end{minipage}
\begin{minipage}[t]{.51\linewidth}
\centering
\begin{tabular}{|c|c|c|ccc|}
\hline
\multirow{2}{*}{Method} & \multirow{2}{*}{Input} & \multirow{2}{*}{RRN} & \multicolumn{3}{c|}{Boundary} \\
 &  &  & \textcolor{orange}{P} & \textcolor{cyan}{R} & \textcolor{purple}{F} \\ \hline
Mask R-CNN & RGB & & 46.7 & 26.3 & 29.5 \\
Mask R-CNN & RGB & \cmark & 52.8 & 28.8 & 33.3 \\
Mask R-CNN & Depth & & 49.6 & 40.3 & 43.1 \\
Mask R-CNN & Depth & \cmark & 69.0 & 55.2 & 59.8 \\
Mask R-CNN & RGB-D & & 57.3 & 52.1 & 53.8 \\
Mask R-CNN & RGB-D & \cmark & 63.4 & 57.0 & 59.2\\
\hline
\end{tabular}
\end{minipage}

\caption{(left) Ablation experiments for UOIS-Net-2D on OSD \cite{richtsfeld2012segmentation}. O/C denotes the Open/Close morphological transform, while CCC denotes Closest Connected Component. (right) Refining Mask R-CNN results with RRN (trained on TOD) on OSD.}
\label{table:ablation}
\end{table*}

\textbf{Comparison to baselines.} 
We compare to baselines shown in \cite{suchi2019easylabel}, which include GCUT \cite{felzenszwalb2004efficient}, SCUT \cite{pham2018scenecut}, LCCP \cite{christoph2014object}, and V4R \cite{potapova2014incremental}. In \cite{suchi2019easylabel}, these methods were only evaluated on the ARID20 and YCB10 subsets of OCID, so we compare our results these subsets as well. These baselines are designed to provide over-segmentations (i.e., they segment the whole scene instead of just the objects of interest). To allow a more fair comparison, we set all predicted masks smaller than 500 pixels to background, and set the largest mask to table label (which is not considered in our metrics). Results are shown in Table \ref{table:baseline_comparison}. Because the baselines aim to over-segment the scene, the precision is in general low while the recall is high. LCCP is designed to segment convex objects (most objects in OCID are convex), but its predicted boundaries are noisy due to utilizing depth only. Recall that OCID labels suffer from depth sensor noise, which explains why LCCP's boundary recall is quite high (see Section~\ref{subsec:qualitative_2D} for a visual example). Both SCUT and V4R utilize models trained on real data, putting them at an advantage with respect to UOIS-Net-2D. V4R was trained on OSD \cite{richtsfeld2012segmentation} which has an extremely similar data distribution to OCID, giving V4R a substantial advantage and making it the strongest baseline in~\cite{suchi2019easylabel}. However, our UOIS-Net-2D (line 5 in Table~\ref{table:baseline_comparison}), despite never having seen any real data, significantly outperforms these baselines on F-measure.

\revisiontext{
\textbf{Comparison to SOTA.}
In Table~\ref{table:input_comparison_sota}, we compare UOIS-Net-2D to two SOTA methods, Mask R-CNN~\cite{he2017mask} and PointGroup~\cite{jiang2020pointgroup}, both trained on RGB-D from TOD. While Mask R-CNN is a general method for detection that can be applied to different types of input modes, the more recent PointGroup requires point clouds and utilizes a sparse convolutional backbone. We see that UOIS-Net-2D (line 6) outperforms Mask R-CNN (line 3) on OSD and slightly on OCID. On the other hand, PointGroup (line 4) provides comparable performance to UOIS-Net-2D. Note, however, that PointGroup reasons in 3D with center voting while UOIS-Net-2D does not. In that sense, Mask R-CNN is more similar to UOIS-Net-2D.
}

\begin{figure*}[t!]
\begin{center}
\includegraphics[width=\linewidth]{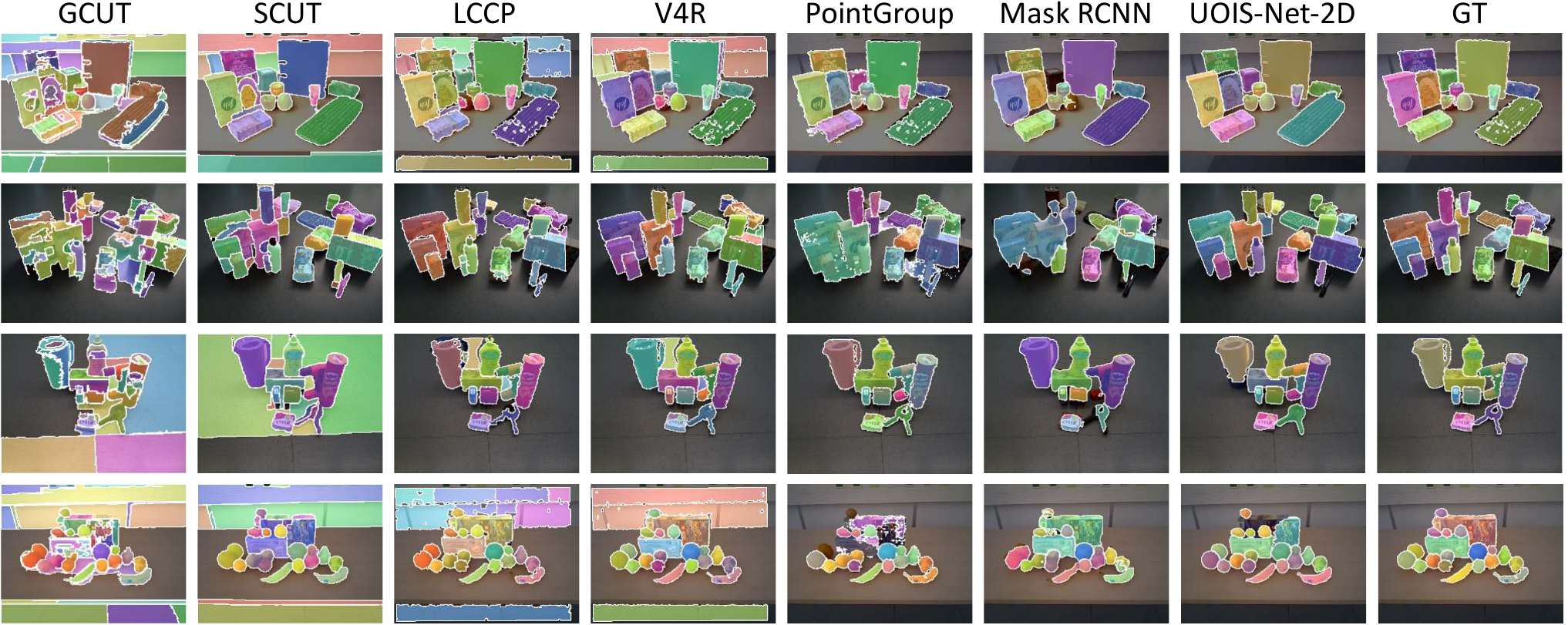}
\caption{Comparison of UOIS-Net-2D with baselines, Mask R-CNN, \revisiontext{and PointGroup} on OCID \cite{suchi2019easylabel}. 
LCCP and V4R operate on depth only, thus are subject to noise from depth sensors. 
However, this is also true for the ground truth segmentation produced by OCID~\cite{suchi2019easylabel}. 
Our proposed UOIS-Net-2D provides sharp and accurate masks in comparison to all of the baselines.
}
\label{fig:qualitative_baseline_comparison}
\end{center}
\end{figure*}

Note that the performance of UOIS-Net-2D is similar to Mask R-CNN \revisiontext{and PointGroup} on OCID in terms of boundary F-measure. This result is misleading: it turns out that using the RRN to refine the initial masks produced by UOIS-Net results in degraded quantitative performance on OCID, while the qualitative results are better. This is due to OCID's noisy label boundaries.
An illustration of this can be found in the first example (row) of Figure~\ref{fig:qualitative_baseline_comparison}. Table~\ref{table:OCID_issue} shows the performance of UOIS-Net without applying RRN (DSN and IMP only) on OCID and OSD. This method utilizes only depth and predicts segmentation boundaries that are aligned with the sensor noise. In this setting, UOIS-Net-2D outperforms \revisiontext{both SOTA methods, with a significant gain in boundary F-measure and a minor gain in overlap F-measure. In fact, UOIS-Net-3D also gains performance on OCID when removing the RRN which further demonstrates the issue with OCID labels.}. On the other hand, OSD has manually annotated labels and this issue is not present, \revisiontext{and our methods' performances deteriorate in the absence of the RRN. In particular, UOIS-Net-2D drops almost 20\% relatively without it}.

\textbf{Effect of input mode.}
To evaluate how different input modes affect results, we train Mask R-CNN on RGB, depth, and RGB-D and compare it to UOIS-Net-2D in Table~\ref{table:input_comparison_sota}. Training Mask R-CNN on synthetic RGB only poorly generalizes from Sim-to-Real. Training on depth drastically boosts this generalization, which is in agreement with \cite{mahler2017dex, danielczuk2018segmenting, seita2018robot}. When training on RGB-D, we posit that Mask R-CNN relies heavily on depth as adding RGB to depth results in little change. However, UOIS-Net-2D exploits RGB and depth separately, leading to better results on OSD while being trained on the exact same synthetic dataset. Furthermore, when our DSN is trained directly on RGB-D (line 4, Table \ref{table:input_comparison_sota}), we see a drop in performance, suggesting that training directly on (non-photorealistic) RGB is not the best way of utilizing synthetic data.

\textbf{Degradation of training on non-photorealistic simulated RGB.}
To quantify how much non-photorealistic RGB degrades performance, we train an RRN on real data. This serves as an approximate upper bound on how well the synthetically-trained RRN can perform. We use the instance masks from OID~\cite{OpenImagesSegmentation} and show results in Table \ref{table:RRN_real_data}. Both models share the same DSN and IMP. The Overlap measures are roughly the same, while the RRN trained on OID has slightly better performance on the Boundary measures. This suggests that while there is still a gap, our method is surprisingly not too far off, considering that we train with non-photorealistic synthetic RGB. We conclude that mask refinement with RGB is an easier task to transfer from Sim-to-Real than directly segmenting from RGB.

\textbf{Ablation studies.}
We report ablation studies on OSD to evaluate each component of our proposed method in Table \ref{table:ablation} (left).
We omit the Overlap P/R/F results since they follow similar trends to Boundary P/R/F. 
Running the RRN on the raw masks output by DSN without the IMP module actually hurts performance as the RRN cannot refine such noisy masks. Adding the open/close morphological transform and/or the closest connect component results in much stronger results, showing that the IMP is crucial in robustifying our proposed method. In these settings, applying the RRN significantly boosts Boundary P/R/F. In fact, Table \ref{table:ablation} (right) shows that applying the RRN to the Mask R-CNN results effectively boosts the Boundary P/R/F for all input modes on OSD, showing the efficacy of the RRN.
Note that even with this refinement, the Mask R-CNN results are outperformed by our method.


\subsection{2D Qualitative Results}
\label{subsec:qualitative_2D}


\textbf{Comparison to baselines and SOTA.} First, we show qualitative results on OCID of baseline methods, Mask R-CNN (trained on RGB-D), PointGroup, and UOIS-Net-2D in Figure \ref{fig:qualitative_baseline_comparison}. It is clear that the baseline methods suffer from over-segmentation issues; they segment the table and background into multiple pieces. This is especially the case for methods that utilize RGB as an input (GCUT and SCUT); the objects are often over-segmented as well. 
In example (row) 1 of Figure~\ref{fig:qualitative_baseline_comparison}, the ground truth label for the keyboard is riddled with holes. Methods that operate only on depth (LCCP and V4R) mirror these noisy boundaries, leading to inflated boundary P/R/F measures.
Despite this, our quantitative results still outperform these baselines.

The main failure mode for Mask R-CNN is that it tends to undersegment objects. This is the typical failure mode of top-down instance segmentation algorithms in clutter. A close inspection of Figure \ref{fig:qualitative_baseline_comparison} shows that Mask R-CNN frequently segments multiple objects as one. Examples 1 and 4 show undersegmentation of many neighboring small objects, and example 2 shows undersegmentation of larger objects. \revisiontext{Since PointGroup requires point clouds, it is susceptible to degradation from depth sensor noise as well. Additionally, some masks (e.g. example 4) can be quite noisy, which we hypothesize is due to the rudimentary breadth-first search clustering algorithm.}

On the other hand, our method utilizes depth and RGB separately to provide sharp and accurate masks. Because UOIS-Net-2D leverages RGB after depth, it can fix the issues with depth sensors shown in example 1. Additionally, our method can segment complicated structures such as stacks (example 2 and 4) and cluttered environments (example 4). 

\begin{figure*}[th!]
\begin{center}
\includegraphics[width=\linewidth]{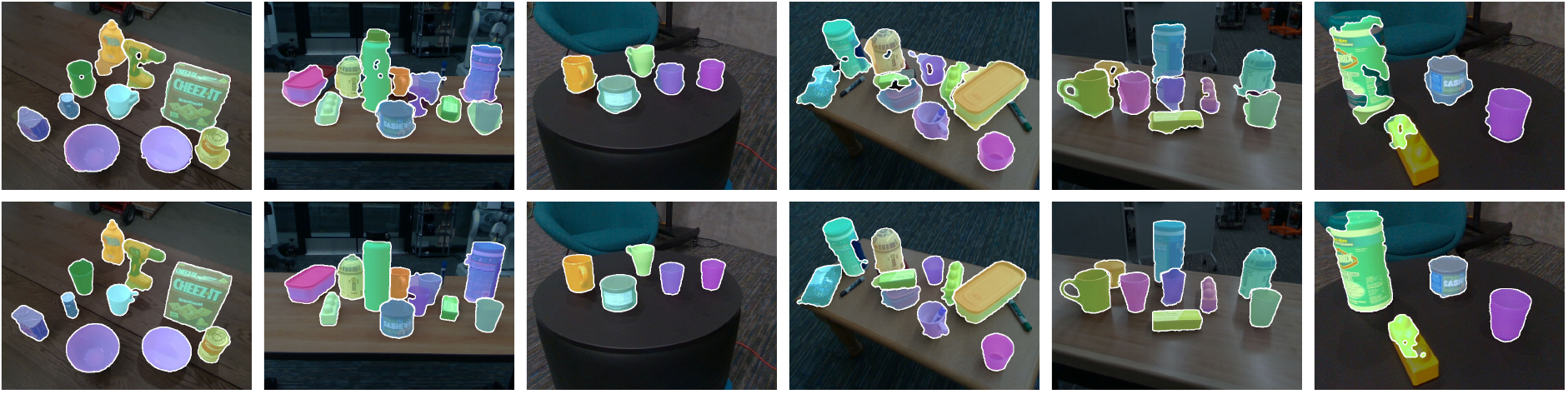}
\caption{Mask refinements with RRN: before (top) refinement (after IMP), and after refinement (bottom). 
}
\label{fig:qualitative_refinement}
\end{center}
\end{figure*}

\textbf{RRN Refinements.}
In Figure \ref{fig:qualitative_refinement}, we qualitatively show the effect of the RRN. The top row shows the masks before refinement (after IMP), and the bottom row shows the refined masks. These images were taken in our lab with an Intel RealSense D435 RGB-D camera to demonstrate the robustness of our method to camera viewpoint variations and distracting backgrounds, as OCID and OSD have relatively simple backgrounds. 
Due to noise in the depth sensor, it is impossible to get sharp and accurate predictions from depth alone without using RGB.
Our RRN can provide sharp masks even when the boundaries of objects are occluding other objects (examples 2 and 5).
Our RRN is able to fix squiggly mask boundaries and patch up large missing chunks in the initial mask.

\begin{figure*}[th!]
\begin{center}
\includegraphics[width=\linewidth]{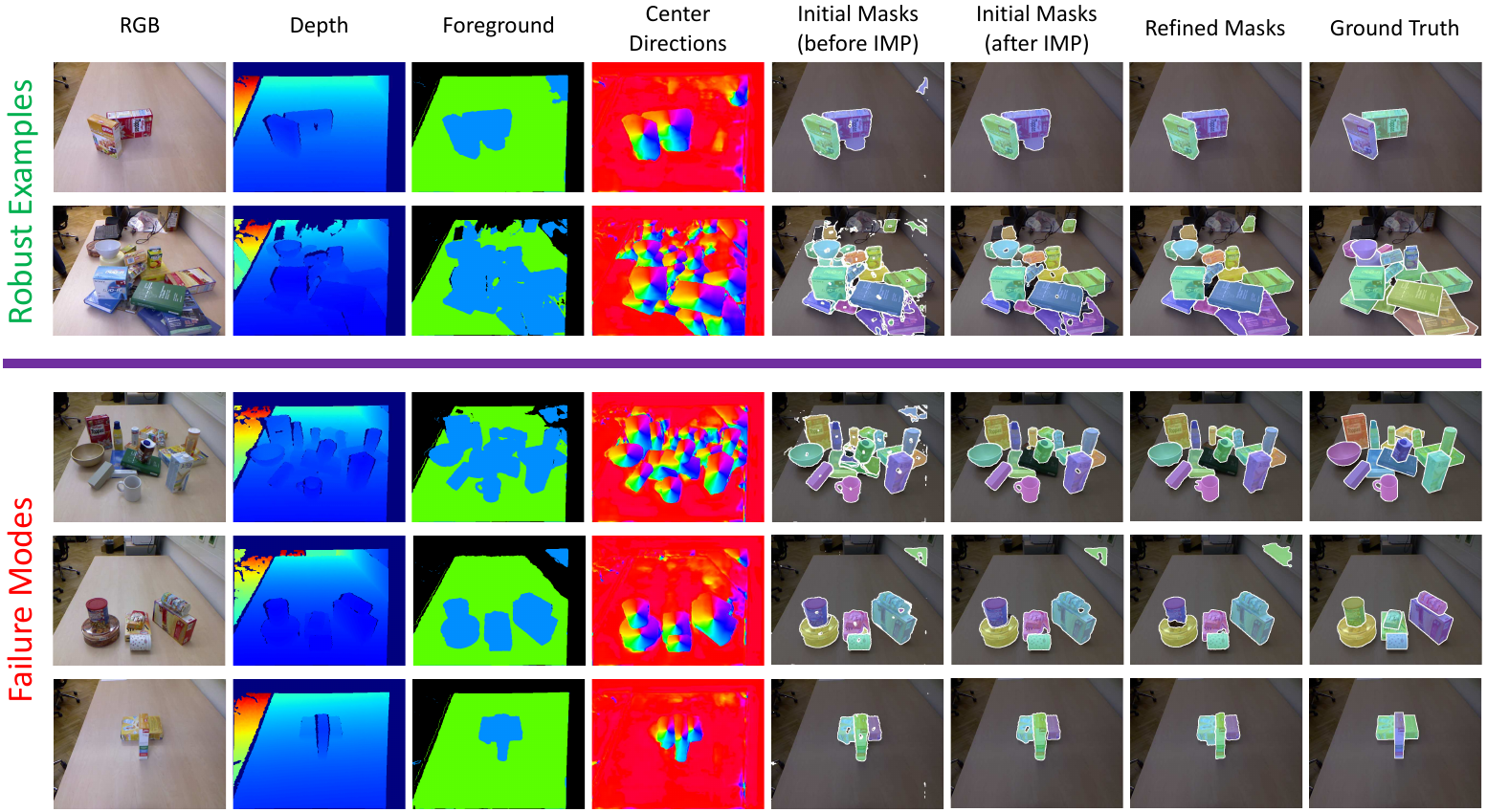}
\caption{
Outputs (and intermediate) of UOIS-Net-2D are visualized. 
We demonstrate the robustness of the Hough voting layer and the IMP (top) and show common failure modes (bottom).
See text for details.
Best viewed in color on a computer screen. 
}
\label{fig:robustness_failure}
\end{center}
\end{figure*}

\textbf{Robustness.} In Figure~\ref{fig:robustness_failure} (top), we demonstrate how our method is robust to errors in the pipeline. 
In row 1, the DSN produces a false positive foreground region in the top right of the image. However, there is not enough evidence in the center directions (not enough discretized angles are present), thus the Hough voting layer suppresses this potential object. 
In row 2, there are many spurious center direction predictions, however these are not considered as potential object centers by the Hough voting layer because they are not detected by the foreground mask. 
Additionally, one can see the many holes and spurious mask components in column 5. 
As seen in the ablation studies in Section~\ref{subsec:quantitative_2D}, applying the RRN here actually degrades the performance.
The IMP cleans this image, making the refinement task easier for the RRN.

\textbf{Failure Modes.}
We show some failure modes of our DSN in Figure \ref{fig:robustness_failure} (bottom) on the OSD dataset. 
In row 1, we see an undetected object (green book) because the center of the mask is occluded. In the center direction image, there is no convergent point for this object which would be considered as the object center with all discretized angles pointing to it. 
In row 2, there is a false positive region detected by the DSN in the top right corner, which cannot be undone by the RRN. Lastly, in row 3, the DSN cannot correctly segment an object whose mask has been split in two by an occluding object.

Our RRN also has some common failure modes. Column 4 in Figure~\ref{fig:qualitative_refinement} shows that the RRN can fail when the object has a complex texture. Column 6 demonstrates that if there is not enough padding to the initial mask to see the entire object, the RRN cannot segment the entire object (lego block). In column 2, the RRN has a tough time fixing the segmentation mask when the DSN has incorrectly undersegmented a few objects together (the two cups to the left of the eraser).

\subsection{3D Quantitative Results}
\label{subsec:quantitative_3D}


\textbf{Comparison to baselines.} We compare the full UOIS-Net-3D with an RRN trained on TOD to all previous baselines from~\cite{suchi2019easylabel} and our UOIS-Net-2D in Table~\ref{table:baseline_comparison}. Discussion of all baselines and UOIS-Net-2D is given in Section~\ref{subsec:quantitative_2D}. Compared to UOIS-Net-2D, the 3D version substantially increases the recall in both the overlap and boundary metrics, leading to a relative increase of 4.5\% in overlap F-measure and 5.5\% in boundary F-measures. 
In fact, UOIS-Net-3D, despite no convexity assumptions, gets quite close to the overlap recall of LCCP, which segments objects based on convexity. 
Note that only the DSN structure is changed when moving from 2D to 3D, thus this extra recall is mainly due to the 3D center voting structure and better initial masks. 

\textbf{Comparison to SOTA.}
In Table~\ref{table:input_comparison_sota}, we show comparisons of UOIS-Net-3D with SOTA methods Mask R-CNN~\cite{he2017mask} \revisiontext{and PointGroup~\cite{jiang2020pointgroup}}, and the previous UOIS-Net-2D on OCID and OSD. Again, the trend of performance increase from UOIS-Net-2D is similar to before, where a significant increase in recall leads to a boost in F-measure. On OCID, UOIS-Net-3D receives a relative increase of 5.8\% in overlap F-measure and a 6.7\% in boundary F-measure. Compared to the best performing Mask R-CNN, our performance provides a relative increase of 8.1\% on overlap F-measure and 6.0\% on boundary F-measure, while being trained on the exact same dataset (TOD). \revisiontext{Additionally, we outperform PointGroup by 7.9\% and 6.3\% on overlap and boundary F-measures, respectively. Note that UOIS-Net-3D performs center voting in a similar fashion to PointGroup, however we believe our novel loss functions target cluttered environments more effectively.} On OSD overlap F-measure, we provide a relative increase of 4.3\% over UOIS-Net-2D, 12.4\% over Mask R-CNN, \revisiontext{and 5.7\% over PointGroup}. For boundary F-measure, we show 8.5\% over UOIS-Net-2D, 32.3\% over Mask R-CNN, and \revisiontext{8.9\% over PointGroup}. Note that while our recall jumps modestly in comparison to UOIS-Net-2D, the precision increase is more sizable.

\begin{table*}[t]
\centering
\begin{tabular}{|cccc|ccc|ccc|ccc|ccc|}
\hline
\multicolumn{4}{|c|}{\multirow{2}{*}{Loss Functions}} & \multicolumn{6}{c|}{OCID \cite{suchi2019easylabel}} & \multicolumn{6}{c|}{OSD \cite{richtsfeld2012segmentation}} \\ \cline{5-16}
&  &  &  & \multicolumn{3}{c|}{Overlap} & \multicolumn{3}{c|}{Boundary} & \multicolumn{3}{c|}{Overlap} & \multicolumn{3}{c|}{Boundary} \\ \hline
$\ell_{fg}$ & $\ell_{co}$ & $\ell_{cl}$ & $\ell_{sep}$ & \textcolor{orange}{P} & \textcolor{cyan}{R} & \textcolor{purple}{F} & \textcolor{orange}{P} & \textcolor{cyan}{R} & \textcolor{purple}{F} & \textcolor{orange}{P} & \textcolor{cyan}{R} & \textcolor{purple}{F} & \textcolor{orange}{P} & \textcolor{cyan}{R} & \textcolor{purple}{F} \\ \hline
\cmark & \cmark &  &  &  
77.9 & 79.8 & 78.6 & 71.2 & 68.0 & 68.9 & 79.4 & 80.1 & 79.8 & 74.5 & 66.5 & 69.8 \\
\cmark & \cmark & \cmark &  &  
77.0 & 81.3 & 78.7 & 71.9 & 68.2 & 69.0 & 72.8 & 74.9 & 72.8 & 67.2 & 61.0 & 62.2 \\
\cmark & \cmark &  & \cmark &  
85.5 & 87.9 & 86.5 & 80.9 & 77.8 & 78.8 & 84.0 & 85.1 & 84.5 & 75.0 & 74.9 & 74.6 \\
\cmark & \cmark & \cmark & \cmark &  
86.0 & 88.2 & 86.9 & 83.1 & 77.6 & 79.9 & 86.0 & 85.1 & 84.8 & 81.0 & 75.3 & 77.3 \\ \hline
\end{tabular}
\caption{Ablation study over loss functions of UOIS-Net-3D. Our novel separation loss $\ell_{sep}$ is crucial to obtaining state-of-the-art performance. Note that we are using an RRN trained on real data.}
\label{table:loss_function_ablation}
\end{table*}

\begin{table*}[t]
\centering

\begin{tabular}{|c|c|ccc|ccc|ccc|ccc|}
\hline
\multirow{3}{*}{\parbox{1.75cm}{\centering DSN reasoning}} & \multirow{3}{*}{\parbox{1.75cm}{\centering ESP Module}} & \multicolumn{6}{c|}{OCID \cite{suchi2019easylabel}} & \multicolumn{6}{c|}{OSD \cite{richtsfeld2012segmentation}} \\ 
\cline{3-14} & & \multicolumn{3}{c|}{Overlap} & \multicolumn{3}{c|}{Boundary} & \multicolumn{3}{c|}{Overlap} & \multicolumn{3}{c|}{Boundary} \\
& & \textcolor{orange}{P} & \textcolor{cyan}{R} & \textcolor{purple}{F} & \textcolor{orange}{P} & \textcolor{cyan}{R} & \textcolor{purple}{F} & \textcolor{orange}{P} & \textcolor{cyan}{R} & \textcolor{purple}{F} & \textcolor{orange}{P} & \textcolor{cyan}{R} & \textcolor{purple}{F} \\ \hline
\multirow{2}{*}{\parbox{1.75cm}{\centering 2D}} & \xmark &
87.9 & 79.6 & 81.7 & 84.0 & 69.1 & 74.1 & 81.2 & 83.3 & 81.7 & 69.8 & 73.7 & 70.8 \\
& \cmark & 
87.6 & 85.0 & 85.7 & 84.4 & 73.4 & 77.8 & 84.6 & 82.3 & 82.8 & 75.5 & 71.6 & 72.4 \\ \hline
\multirow{2}{*}{\parbox{1.75cm}{\centering 3D}} & \xmark &
83.9 & 85.6 & 84.4 & 79.1 & 76.4 & 77.2 & 85.1 & 85.7 & 85.1 & 78.9 & 76.5 & 77.3 \\
& \cmark & 
86.0 & 88.2 & 86.9 & 83.1 & 77.6 & 79.9 & 86.0 & 85.1 & 84.8 & 81.0 & 75.3 & 77.3 \\
\hline

\end{tabular}
\caption{Ablation study to test the significance of a wider receptive field. Using the ESP module~\cite{mehta2018espnet} in the DSN architecture improves performance for both UOIS-Net-2D and UOIS-Net-3D. Note that we are using an RRN trained on real data.}
\label{table:ESP_module_ablation}
\end{table*}

\begin{figure}[t]
\begin{center}
\includegraphics[width=\linewidth]{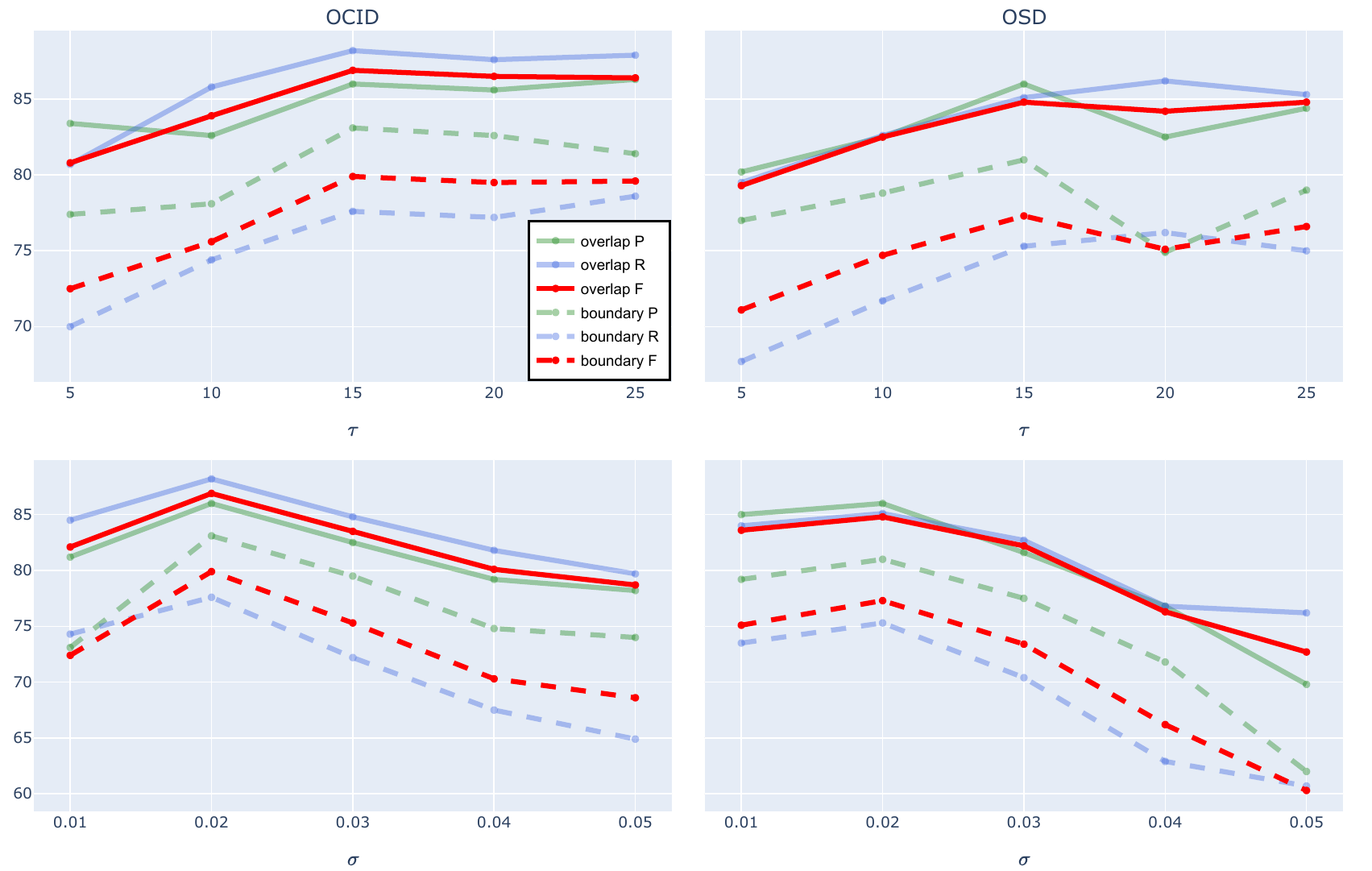}
\caption{Ablation study to test the sensitivity of UOIS-Net-3D with respect to $\tau, \sigma$. 
Best viewed by zooming in on a computer.}
\label{fig:tau_sigma_ablation}
\end{center}
\end{figure}

\textbf{RRN ablation.} We refer readers to Table~\ref{table:RRN_real_data} to show the full power of our method. When using an RRN trained on \revisiontext{OID}~\cite{OpenImagesSegmentation}, our full UOIS-Net-3D further increases the boundary F-measures, leading to 79.9 points on OCID and 77.3 points on OSD, which is 7.8\% and 9.2\% higher than the full UOIS-Net-2D on OCID and OSD, respectively. In the rest of this section, all UOIS-Net-3D's will use an RRN trained on OID in order to show the full performance of our method. 

\begin{figure*}[th!]
\begin{center}
\includegraphics[width=\linewidth]{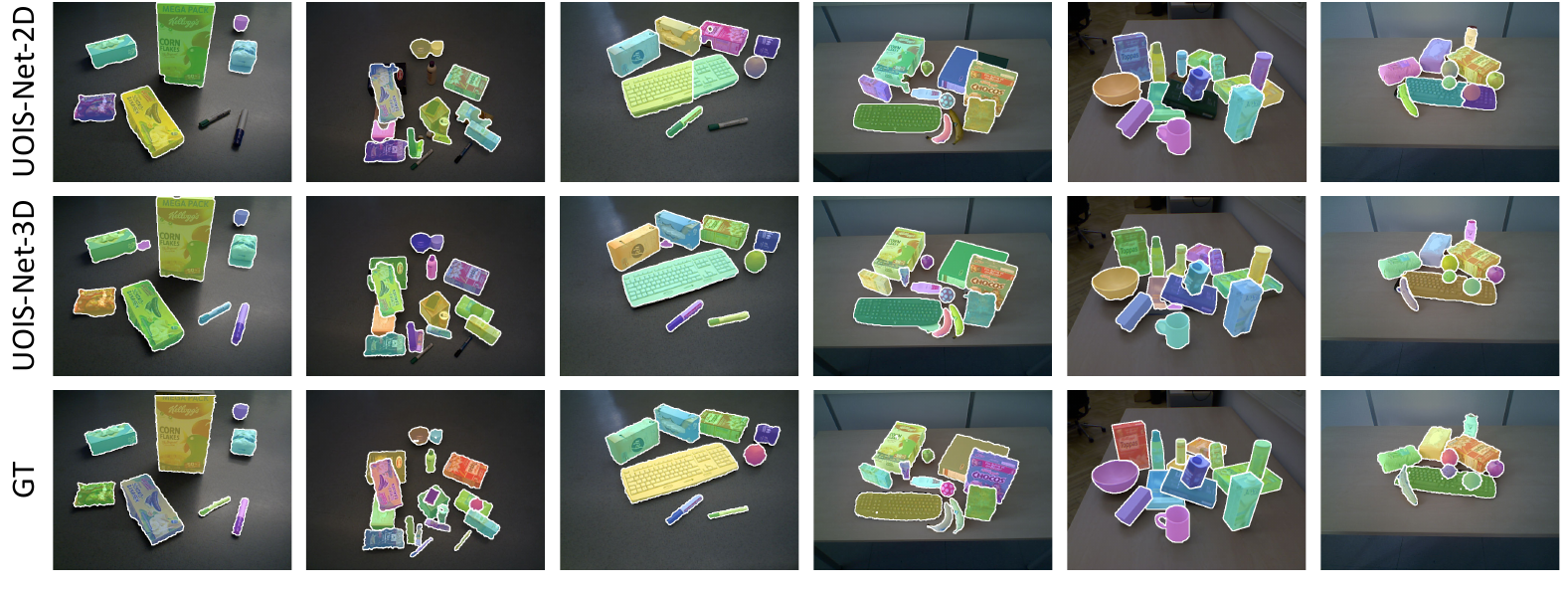}
\caption{Qualitative comparison of UOIS-Net-2D to UOIS-Net-3D on OCID. 
}
\label{fig:qualitative_2D_to_3D_comparison}
\end{center}
\end{figure*}

\textbf{Loss function ablation.}
At the minimum, we require $\ell_{fg}$ and $\ell_{co}$ to train UOIS-Net-3D. In Table~\ref{table:loss_function_ablation}, UOIS-Net-3D actually performs poorer than UOIS-Net-2D (see Table~\ref{table:RRN_real_data}) with these two losses only. When adding the cluster loss $\ell_{cl}$, we get roughly the same performance on OCID. Our intuition is that this loss encourages the same behavior as $\ell_{co}$, namely that the points should cluster near the ground truth object center, thus it doesn't introduce anything new. However, adding this loss significantly hurts performance on OSD; we posit that this result is noisy due to the small size of OSD. In the third row, we show that adding our novel $\ell_{sep}$ loss relatively boosts overlap F-measure by 10.1\% and boundary F-measure by 14.4\% compared to the base model ($\lambda_{fg}\ell_{fg} + \lambda_{co}\ell_{co}$), demonstrating it is crucial for obtaining good performance. 
$\ell_{sep}$ pushes center votes away from center votes of other objects in order to make the post-processing GMS clustering easier. Note that the center votes don't necessarily need to be close to the ground truth object centers; they simply need to be easily clustered, and this loss allows for that. This property is crucial to getting strong performance in clutter as the numbers demonstrate. We will also visually illustrate this phenomena in Section~\ref{subsec:qualitative_3D}. 
Lastly, since the object clusters are not necessarily near the ground truth object centers, $\ell_{cl}$ now encourages something that $\ell_{co}$ does not: it encourages that the center votes are tightly packed wherever they are placed, further easing the job of the post-processing GMS clustering. In line 4 we see it provides some extra performance gain.

\textbf{ESP module ablation.}
We evaluate the significance of using the ESP module~\cite{mehta2018espnet} to obtain a higher receptive field by testing both UOIS-Net-3D and UOIS-Net-2D (which normally does not include the ESP module) with and without the ESP modules embedded into the DSN in Table~\ref{table:ESP_module_ablation}. We immediately see that obtaining a higher receptive field is beneficial to the performance. For UOIS-Net-3D, we see relative gains on OCID of 3.0\% and 3.5\% on overlap and boundary F-measures, respectively. On OSD, we see similar performance. For UOIS-Net-2D, we see a large relative gain on OCID of 4.9\% and 5\% on overlap and boundary F-measures, respectively. Note that the DSN with ESP modules has slightly fewer parameters than without them, thus these experiments highlight the usefulness of a higher receptive field.

\textbf{$\tau$ ablation.}
In the top row of Figure~\ref{fig:tau_sigma_ablation}, we test the sensitivity of our model to the settings of $\tau$ over $[5, 10, 15, 20, 25]$, which is used in $\ell_{sep}$. We visualize both overlap and boundary P/R/F measures. Note that as $\tau$ is essentially a multiplicative factor on the Euclidean distance in Eq. ($\ref{eq:sep_loss}$); thus, as $\tau$ becomes larger, the Euclidean distance becomes more inflated and objects do not need to be pushed as far in order to minimize $\ell_{sep}$. It is clear to see Figure~\ref{fig:tau_sigma_ablation} that all metrics increases as $\tau$ increases, and they plateau after $\tau$ hits a high enough value, in this case 15. This suggests that we only need a small level of encouragement to push the object centers away from each other in order to get strong performance, and that setting $\tau$ too low can degrade performance, potentially by making $\ell_{sep}$ too difficult to minimize.

\textbf{$\sigma$ ablation.}
In the bottom row of Figure~\ref{fig:tau_sigma_ablation}, we test the sensitivity of UOIS-Net-3D to $\sigma$ over $[0.01, 0.02, 0.03, 0.04, 0.05]$, which is used in $\ell_{cl}$. The value of $\sigma$ determines how tightly packed the center votes of an object need to be in order to minimize $\ell_{cl}$. Additionally, it determines how much spread the center votes can have and still be clustered together during the post-processing GMS clustering step. When $\sigma$ is too small, the DSN typically oversegments the objects. On the other hand, when $\sigma$ is too large, center votes from multiple objects get clustered together and undersegmentation occurs. Thus, $\sigma$ should be chosen to reflect the level of clutter in the scenes, i.e. how close the objects are. Our ablation study hints at this idea: the bottom row of Figure~\ref{fig:tau_sigma_ablation} that $\sigma=0.02$ works the best for both OCID~\cite{suchi2019easylabel} and OSD~\cite{richtsfeld2012segmentation}, as they are similar in distribution. In fact, the trends of the graph in both datasets are also similar.
Note that we tried learning $\sigma$ as a parameter of the network, however these experiments did not work well.

\subsection{3D Qualitative Results}
\label{subsec:qualitative_3D}


\begin{figure}[th!]
\begin{center}
\includegraphics[width=\linewidth]{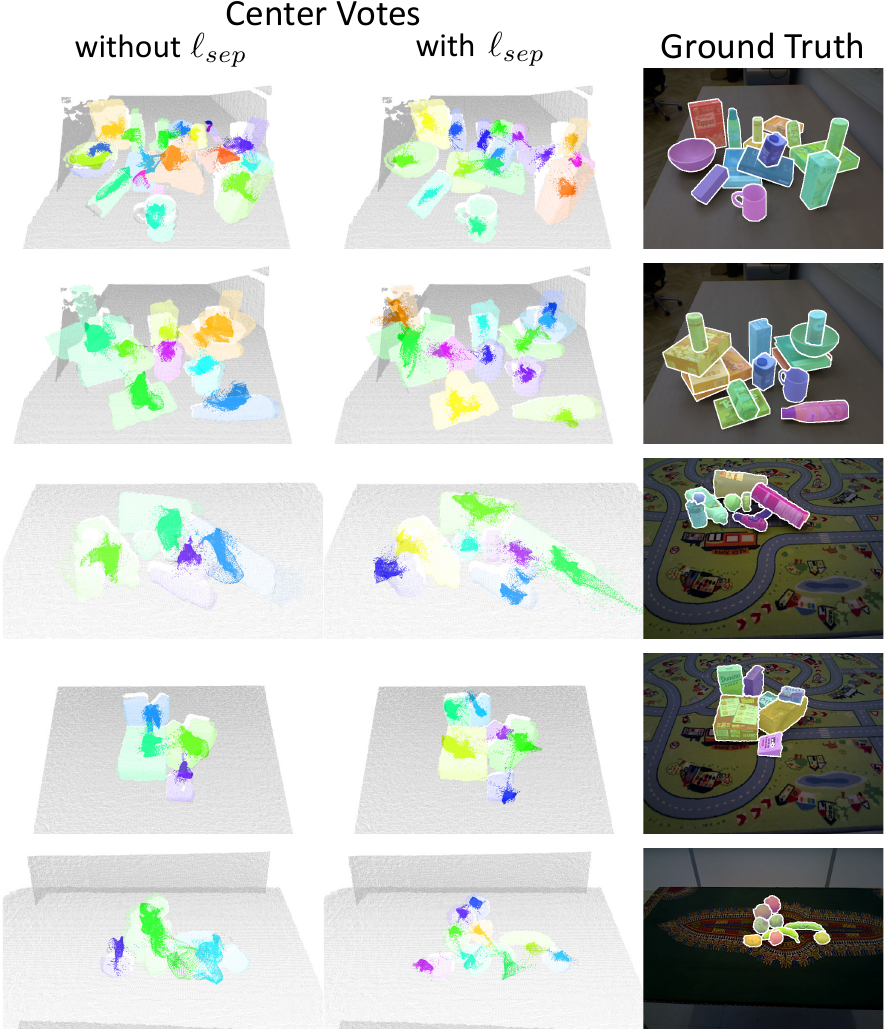}
\caption{Effect of $\ell_{sep}$ on center votes. Columns 1 and 2 show the point cloud (visualized \revisiontext{with Open3D} \cite{Zhou2018}) and center votes overlaid on the image, which are color-coded according to their instance ID.
Best viewed in color and zoomed in.}
\label{fig:qualitative_separation_loss_comparison}
\end{center}
\end{figure}

\textbf{2D vs. 3D comparison.} In Figure~\ref{fig:qualitative_2D_to_3D_comparison}, we qualitatively compare the predictions of UOIS-Net-2D and UOIS-Net-3D to understand how reasoning in 3D can solve the 2D issues mentioned beforehand (in Section~\ref{subsec:qualitative_2D}). The first row of Figure~\ref{fig:robustness_failure} (bottom) exhibits the 2D issue of false negative detection when the 2D center of the object is occluded. However, this is easily corrected when voting for 3D centers, as seen in columns 2 and 5 of Figure~\ref{fig:qualitative_2D_to_3D_comparison} 
In columns 1, 3, and 4, we see that UOIS-Net-3D detects and segments more small, thin objects such as pens and bananas. The 2D Hough voting procedure typically fails to detect enough discretized directions for objects with such shape. This issue is ameliorated when reasoning in 3D, as there are no approximations made with discretized directions. Lastly, in columns 3, 4, and 6, we show that the 3D DSN architecture performs better due to a higher receptive field provided by the ESP modules. 

\textbf{Center Votes.} As shown in Table~\ref{table:loss_function_ablation}, $\ell_{sep}$ is crucial to ensuring UOIS-Net-3D has strong performance. We visually examine the reason for this in Figure~\ref{fig:qualitative_separation_loss_comparison}. In the first and second columns, we visualize the point cloud of the scene with the predicted center votes overlaid on the same image. Both the point cloud and center votes are color coded with respect to their instance segmentation masks. 
It is clear to see that when training without $\ell_{sep}$, the center votes are more spread out and messy, which makes the post-processing GMS clustering more difficult. On the other hand, the full model has much more tightly packed and separated center votes, which leads to better performance. 
Looking at example (row) 5, there are many small objects such as fruits and bananas which are close to each other in spatial proximity. UOIS-Net-3D trained without $\ell_{sep}$ shows that the center votes are mostly jumbled together, while the full model nicely separates the center votes which leads to almost perfect segmentation.
Note that the center votes do not need to be at the center of the object, as long as they can be clustered correctly. In the upper right of example 2, the cylinder and the bowl have very close 3D centers. The 3D DSN predicts the center votes in a way that pushes them apart to ease the post-processing clustering and allows (the full) UOIS-Net-3D to correctly segment these objects.

\begin{figure*}[th!]
\begin{center}
\includegraphics[width=\linewidth]{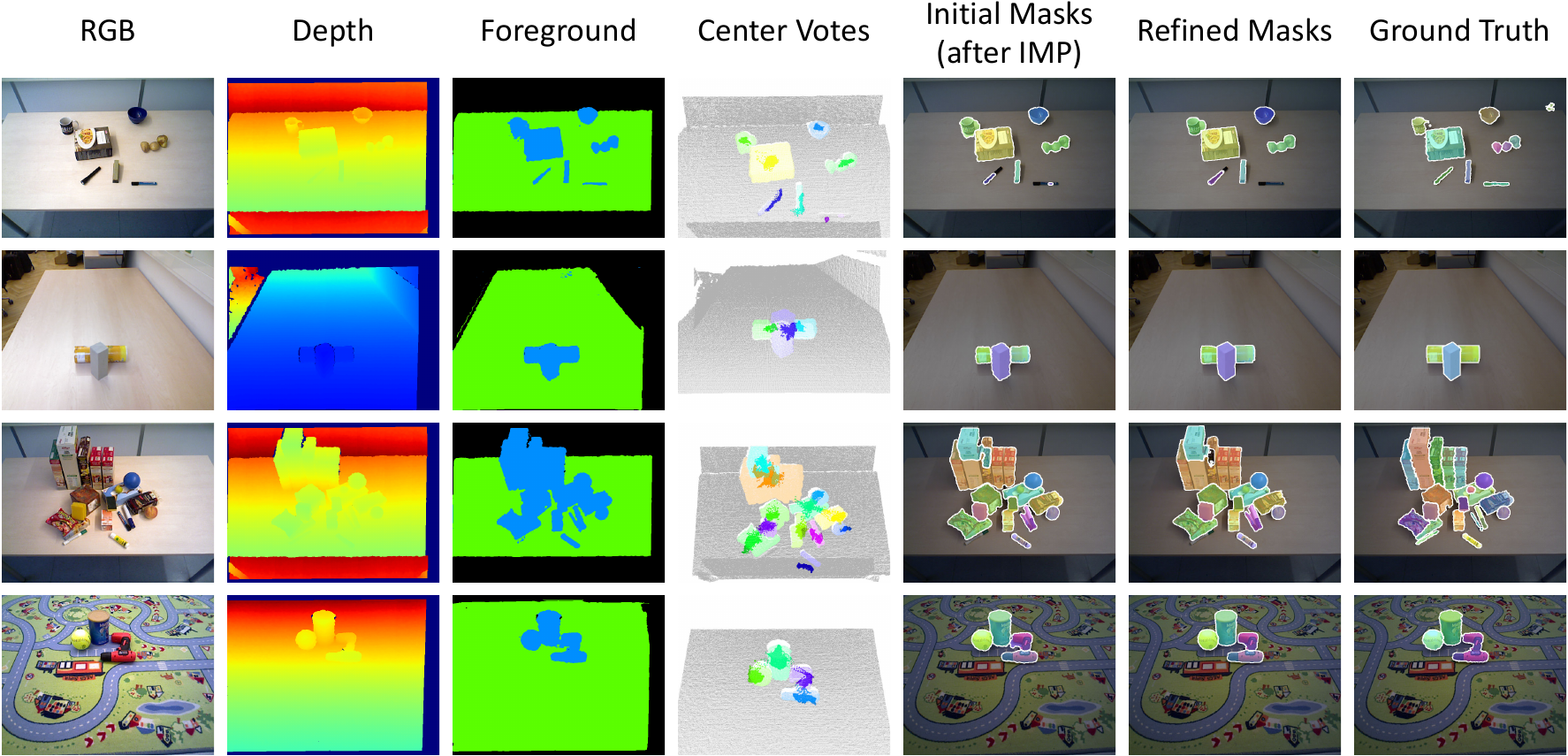}
\caption{Common failure modes of UOIS-Net-3D. See text for details.
Best viewed in color and zoomed in.}
\label{fig:qualitative_3D_failure_modes}
\end{center}
\end{figure*}

\textbf{Failure modes.} In Figure~\ref{fig:qualitative_3D_failure_modes}, we demonstrate common failure modes of UOIS-Net-3D. Row 1 shows that when objects are close together, they may under-segmented into a single segment. The center votes shows that the three small fruits have center votes that are too close to separate in the post-processing clustering step. Another case of under-segmentation is shown in row 3, where multiple objects (cereal boxes) with flat surfaces are lined up such that the depth map shows a large flat surface. It is difficult to correctly separate these boxes from depth alone; appearance is key in providing the correct segmentation in such situations. Row 4 shows that highly nonconvex objects such as power drills can be over-segmented into pieces. The center votes image clearly shows that the DSN believes this object to be two objects. Lastly, row 2 indicates that the 2D issue of attempting to segment a mask that is split into multiple pieces by an occluding object is still present when reasoning in 3D.



\revisiontext{
\subsection{Quantifying Generalization from Sim to Real}
}

\begin{table}[H]
\centering
\begin{tabular}{|c|ccc|ccc|}
\hline
\multirow{2}{*}{Method} & \multicolumn{3}{c|}{Overlap} & \multicolumn{3}{c|}{Boundary} \\
& \textcolor{orange}{P} & \textcolor{cyan}{R} & \textcolor{purple}{F} & \textcolor{orange}{P} & \textcolor{cyan}{R} & \textcolor{purple}{F} \\ \hline
UOIS-Net-2D & 94.3 & 89.0 & 90.9 & 89.7 & 80.6 & 84.1 \\
UOIS-Net-3D & 92.1 & 92.1 & 91.9 & 85.0 & 87.1 & 85.5 \\
Mask R-CNN & 95.4 & 95.2 & 95.1 & 92.3 & 90.0 & 90.9 \\
PointGroup & 97.7 & 95.1 & 96.1 & 93.3 & 89.8 & 91.1 \\
\hline
\end{tabular}
\caption{\revisiontext{Performance on TOD test set (20k images)}.}
\label{table:TOD_test}
\end{table}

\revisiontext{
In order to quantify generalization from simulation to the real-world, we show performance of our methods and SOTA methods Mask R-CNN and PointGroup on the TOD test set. This dataset of 20k images was generated using instances that are not present in the training set. In Table~\ref{table:TOD_test}, we show that both Mask R-CNN and PointGroup outperform UOIS-Net on all metrics on this test set. However, as seen in Table~\ref{table:input_comparison_sota}, UOIS-Net outperforms Mask R-CNN and PointGroup on real-world data, indicating that our methods handle the distribution shift better, which is ultimately what we care about.
}

\subsection{Application in Grasping Unknown Objects}

\begin{figure}[t!]
\begin{center}
\includegraphics[width=\linewidth]{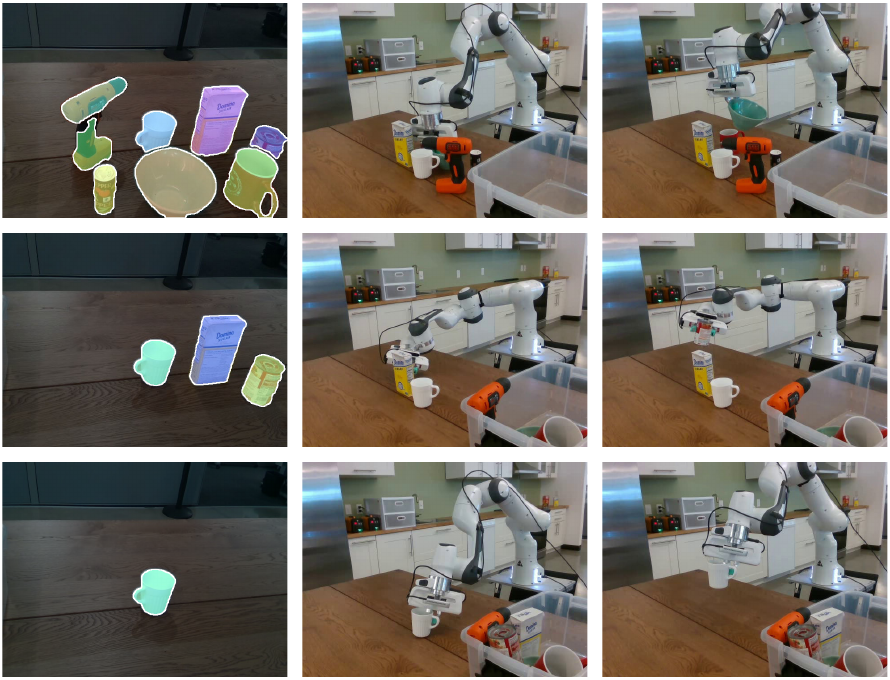}
\caption{Visualization of clearing table using our instance segmentation and 6-DOF GraspNet~\cite{mousavian2019grasp}.}
\label{fig:robot}
\end{center}
\end{figure}

We use our model to demonstrate manipulation of unknown objects in a cluttered environment using a Franka robot with panda gripper and wrist-mounted RGB-D camera. The task is to collect objects from a table and put them in a bin. Object instances are segmented using our method and the point cloud of the closest object to the camera is fed to 6-DOF GraspNet \cite{mousavian2019grasp,murali2020clutteredgrasping} to generate diverse grasps, with other objects considered obstacles. Other objects are represented as obstacles by sampling fixed number of points using farthest point sampling from their corresponding point cloud. The grasp that has the maximum score and has a feasible path is chosen to execute. Figure~\ref{fig:robot} shows the instance segmentation at different stages of the task and also the execution of the robot grasps.
Video of the experiments can be found at the project website\footnote{\url{https://rse-lab.cs.washington.edu/projects/unseen-object-instance-segmentation/}}.
Our method segments the objects correctly most of the time but fails sometimes, such as the over-segmentation of the drill in the scene. Our method considers the top of the drill as one object and the handle as an obstacle. This is because the object is highly nonconvex as discussed in the previous section. 
We conducted the experiment to collect 51 objects from 9 different scenes. Each object is considered to be successfully grasped if the robot can pick it up in maximum 2 attempts. Otherwise, we count that object as failure case and remove it from the scene manually and proceed to other objects. In our trials, the robot successfully grasped 41/51 objects ($80.3\%$ success rate). The failures stem from either imperfections in segmentation or inaccurate generated grasps.
Note that neither our method nor 6-DOF GraspNet~\cite{mousavian2019grasp} are trained on real data.



%% file: conclusion.tex
\section{Conclusion}

We proposed a deep network, UOIS-Net, that separately leverages RGB and depth to provide sharp and accurate masks for unseen object instance segmentation. Our two-stage framework produces rough initial masks using only depth by regressing center votes in either 2D or 3D, then refines those masks with RGB. Surprisingly, our RRN can be trained on non-photorealistic RGB and generalize quite well to real world images. We demonstrated the efficacy of our approach on multiple tabletop environment datasets and showed that our model can provide strong results for unseen object instance segmentation. Finally, we also addressed the weaknesses of our method which will serve as the base of future work.

